\RequirePackage{fix-cm}
\documentclass[twocolumn]{svjour3}
\smartqed 
\usepackage{graphicx}
\usepackage{natbib}
\usepackage{amsmath,amsfonts}
\usepackage{booktabs}
\usepackage{wrapfig}
\usepackage{multirow}
\usepackage{graphicx}
\usepackage{subcaption}
\usepackage{url}

\newcommand{\humanEmoji}{\includegraphics[height=.8em,trim=0 2em 0em 0]{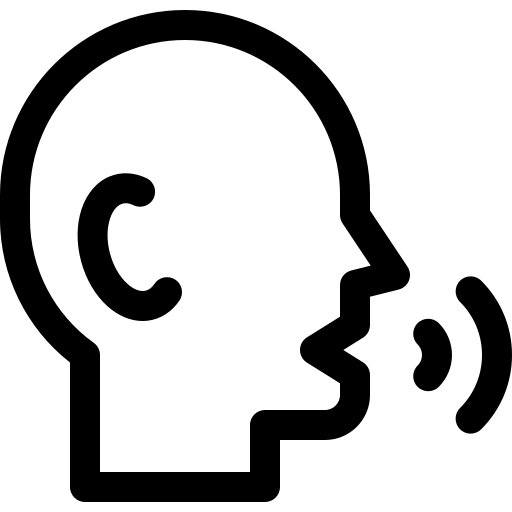}}
\newcommand{\skiEmoji}{\includegraphics[height=.8em,trim=0 2em 0em 0]{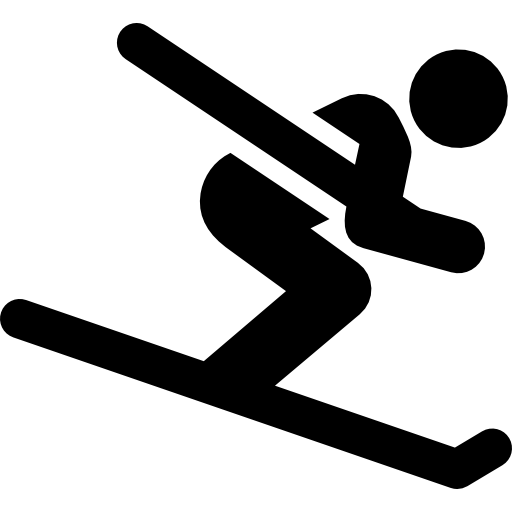}}

\newcommand{\elkEmoji}{\includegraphics[height=.8em,trim=0 2em 0em 0]{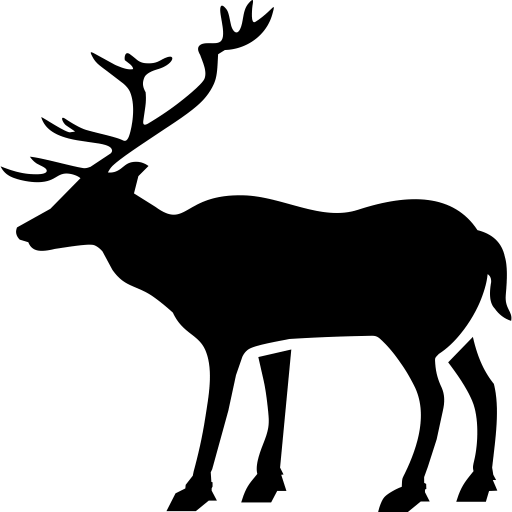}}
\newcommand{\windEmoji}{\includegraphics[height=.8em,trim=0 2em 0em 0]{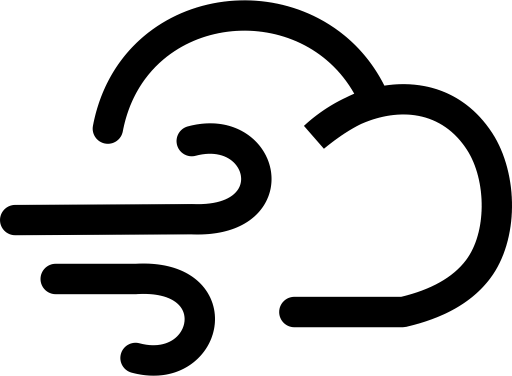}}
\newcommand{\tornadoEmoji}{\includegraphics[height=.8em,trim=0 2em 0em 0]{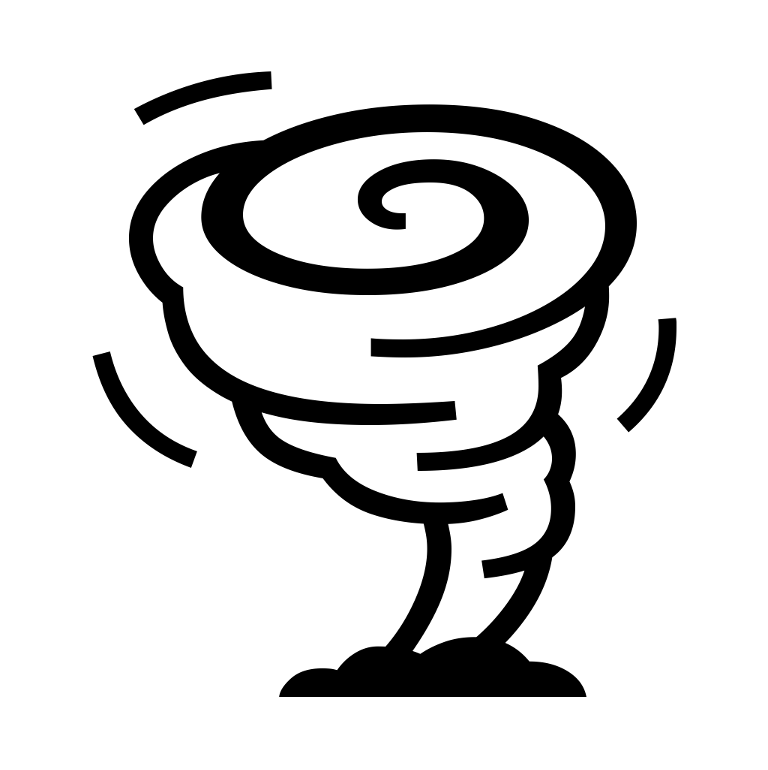}}
\newcommand{\treeEmoji}{\includegraphics[height=.8em,trim=0 2em 0em 0]{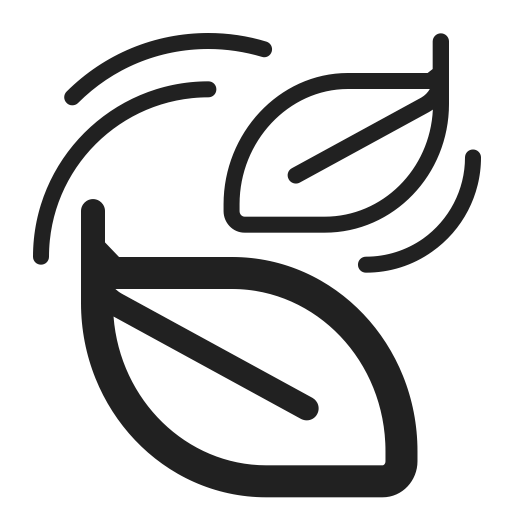}}
\newcommand{\waterEmoji}{\includegraphics[height=.8em,trim=0 2em 0em 0]{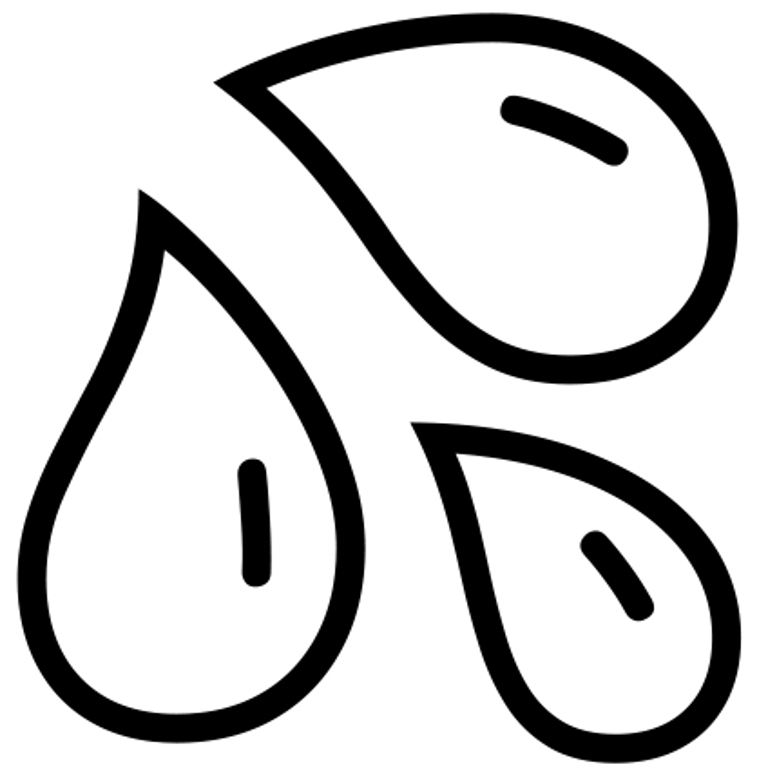}}
\newcommand{\snowEmoji}{\includegraphics[height=.8em,trim=0 2em 0em 0]
{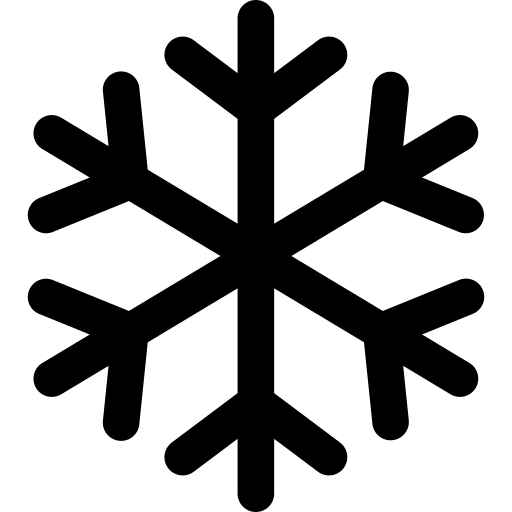}}
\usepackage{xspace}
\def\onedot{.\@\xspace}
\usepackage{mathtools}
\DeclarePairedDelimiter{\norm}{\lVert}{\rVert}
\newcommand{\blank}[1]{\hspace*{#1}}
\newcommand{\Eref}[1]{Eq.~(\ref{#1})}
\newcommand{\ba}{{\mathbf{a}}}
\newcommand{\bb}{{\mathbf{b}}}
\usepackage{enumitem}
\usepackage{multirow}
\usepackage{wrapfig}
\usepackage{colortbl}
\usepackage[normalem]{ulem}
\usepackage{comment}
\usepackage{microtype}
\renewcommand{\paragraph}[1]{\vspace{0.5mm}\noindent\textbf{#1.}\,\,}
\begin{document}

\title{Sound2Vision: Generating Diverse Visuals from Audio through Cross-Modal Latent Alignment}

\author{Kim Sung-Bin         \and
        Arda Senocak        \and
        Hyunwoo Ha        \and
        Tae-Hyun Oh       
}

\institute{Kim Sung-Bin \at
              Department of Electrical Engineering, POSTECH, Pohang, Republic of Korea \\
              \email{sungbin@postech.ac.kr}
           \and
           Arda Senocak \at
              School of Electrical Engineering, KAIST, Daejeon, Republic of Korea \\
              \email{arda.senocak@gmail.com}
            \and
           Hyunwoo Ha \at
              Department of Electrical Engineering, POSTECH, Pohang, Republic of Korea \\
              \email{hyunwooha@postech.ac.kr}
           \and
           Tae-Hyun Oh \at
              Department of Electrical Engineering and Graduate School of Artificial Intelligence, POSTECH, Pohang, Republic of Korea \\
              Institute for Convergence Research and Education in Advanced Technology, Yonsei University, Seoul, Republic of Korea \\
              \email{taehyun@postech.ac.kr}
}

\date{Received: date / Accepted: date}

\maketitle

\begin{abstract}
How does audio describe the world around us? 
In this work, we propose a method for generating images of visual scenes from diverse in-the-wild sounds.
This cross-modal generation task is challenging due to the significant information gap between auditory and visual signals. 
We address this challenge by designing a model that aligns audio-visual modalities by enriching audio features with visual information and translating them into the visual latent space. 
These features are then fed into the pre-trained image generator to produce images.
To enhance image quality, we use sound source localization to select audio-visual pairs with strong cross-modal correlations. 
Our method achieves substantially better results on the VEGAS and VGGSound datasets compared to previous work and demonstrates control over the generation process through simple manipulations to the input waveform or latent space. 
Furthermore, we analyze the geometric properties of the learned embedding space and demonstrate that our learning approach effectively aligns audio-visual signals for cross-modal generation. 
Based on this analysis, we show that our method is agnostic to specific design choices, showing its generalizability by integrating various model architectures and different types of audio-visual data. 
\end{abstract}

\section{Introduction}\label{intro}
Humans possess the unique ability to link sounds to specific visual scenes. For instance, the sounds of birds chirping and branches rustling evoke images of a dense forest, while the sound of water flowing brings to mind a river. These associations between sound and visual scenes also enable humans to infer information, such as the size and distance of sound sources, as well as the presence of objects that are not immediately visible.

\begin{figure*}[tp]
    \centering
    \includegraphics[width=\linewidth]{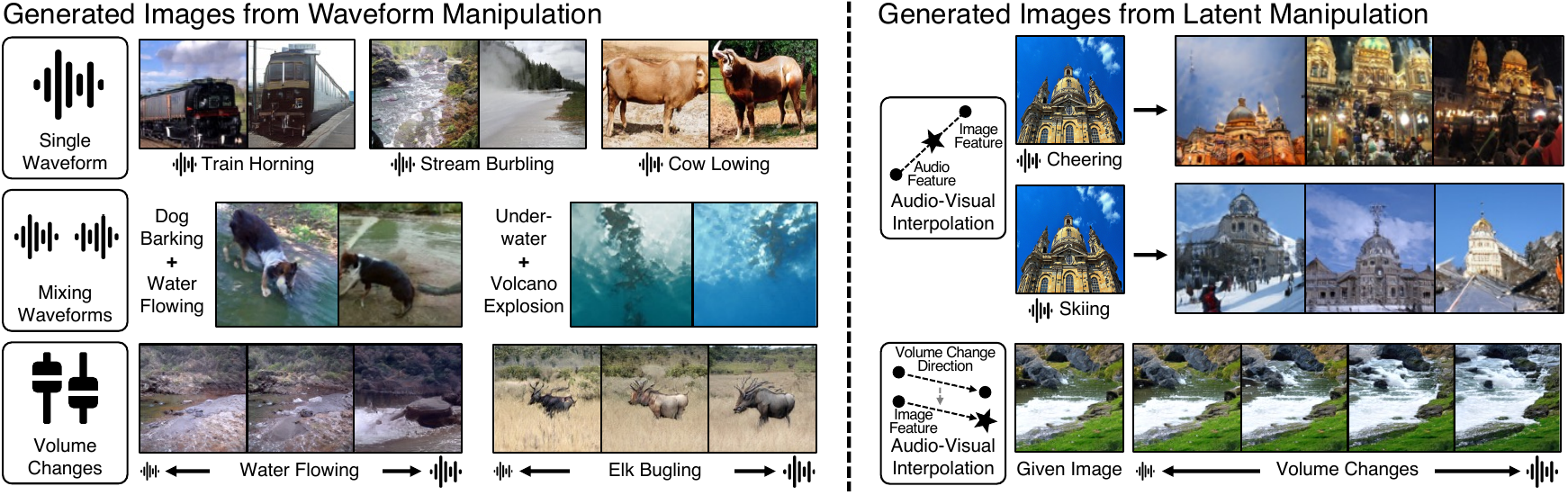}
    \vspace{-5mm}
    \caption{{\bf Sound-to-image generation.} We propose a model that synthesizes images of natural scenes from the sound. Our model is trained solely from paired audio-visual data, without labels or language supervision. Our model's predictions can be controlled by applying simple manipulations to the input waveforms (left), such as by mixing two sounds together or by adjusting the volume. We can also control our model's outputs in latent space, such as by interpolating in directions specified by sound (right).}
    \vspace{-2mm}
    \label{fig:teaser}
\end{figure*}

Recent efforts have been focused on developing multimodal learning systems capable of making such cross-modal predictions by generating visual content based on audio inputs~\citep{soundguide, li2022learning,towards_a2s,s2b,chen2017,cmcgan,s2i}. Nonetheless, these existing approaches face several challenges, including their restriction to simple datasets where there is a strong correlation between images and sounds~\citep{towards_a2s,s2b}, dependence on vision-and-language supervision~\citep{soundguide}, and limitations to only manipulating the style of existing images rather than generating new ones~\citep{li2022learning}.

Overcoming these limitations involves addressing several challenges. Firstly, there is a considerable gap between the modalities of vision and sound, as sound often lacks crucial visual details, such as the shape, color, or spatial positioning of objects. Secondly, the correlation between these modalities can often be inconsistent, \emph{e.g.}, highly contingent or off-sync in timing. 
Cows, for example, only rarely moo, so associating images of cows with ``moo'' sounds requires capturing training examples with the rare moments when on-screen cows vocalize.

In this work, we introduce Sound2Vision, a novel sound-to-image generation model and training method that overcomes existing limitations and can be trained using diverse, unlabeled in-the-wild videos. Initially, we utilize a self-supervised pre-trained image encoder and develop an image generation model that synthesizes images based on the visual features from the encoder. Next, we design an audio encoder that converts input sounds into their corresponding visual features, by training to align the audio with the visual domain. This enables us to synthesize diverse images from sound by converting audio into visual features and then generating an image. To effectively learn such cross-modal generation from complex real-world videos, we employ a sound source localization technique~\citep{less} to select moments with strong cross-modal correlations in the videos and use those moments for training the model.

Our proposed model, Sound2Vision, is evaluated using the VEGAS~\citep{vegas} and VGGSound~\citep{vggsound} datasets. Sound2Vision is capable of synthesizing diverse visual scenes from a wide range of audio inputs, compared to existing models. Additionally, it offers an intuitive method for controlling the image generation process through manipulations at both the input and latent space levels, such as mixing multiple input waveforms, modifying their volumes, and interpolating the features in the latent space, as shown in Fig.~\ref{fig:teaser}.

We further analyze the geometric properties of the learned audio-visual aligned space to understand the effectiveness of our training approach. 
Through an analysis of the modality gap in multi-modal contrastive representation, we empirically demonstrate that our method achieves cross-modal transferability~\citep{zhang2023diagnosing}, which is the core motivation of our proposed approach, \emph{i.e.}, enabling cross-modal generation from audio to sound. 
Building on this analysis, we show that our approach is not dependent on specific design choices and has broad applicability. 
We demonstrate that our training method can adapt to various architectural choices, ranging from GANs to the Latent Diffusion Model (LDM)~\citep{stable}. Additionally, it can be applied to different types of audio-visual datasets, including generic audio-visual data as well as speech and face data~\citep{zhu2022celebv}.

Our main contributions are summarized as follows:
\begin{itemize}
    \item Proposing a 
    sound-to-image generation method that can generate visually rich images from diverse categories of in-the-wild audio in a self-supervised way.
    \item Demonstrating that the samples generated by our model can be controlled by intuitive manipulations in the waveform space in addition to latent space. 
    \item  Showing the effectiveness of training sound-to-image generation using highly correlated audio-visual pairs.
    \item Providing an analysis of the geometric properties of the learned space that enables cross-modal generation.
    \item Demonstrating the generalizability of the training method by successfully incorporating different model architectures and types of audio-visual data.
\end{itemize}

\section{Related work}\label{related}
\paragraph{Audio-visual generation}
The field of audio-visual cross-modal generation is explored through two main directions: vision-to-sound generation and sound-to-vision generation. 
The former, vision-to-sound, has received significant attention, particularly in creating music from images or videos of instruments playing~\citep{audeo,chen2017,cmcgan}, as well as generating more general sounds, such as impact noises from silent videos~\citep{visualsound} and open-domain sounds from images and videos~\citep{tamingsound, regnet, vegas}. Recent work~\citep{luo2024diff,xing2024seeing,zhang2024foleycrafter} has also shown success in generating temporally synchronized audio from generic open-domain silent videos.

Conversely, early works in sound-to-vision primarily focused on specific audio categories, such as musical instruments~\citep{chen2017,cmcgan,strummingbeat,sound2sight}, bird sounds~\citep{s2b}, and human speech~\citep{speech2face}. 
More recent efforts by Wan~\emph{et al}\onedot~\citep{towards_a2s} and Fanzeres~\emph{et al}\onedot~\citep{s2i} have expanded the scope, aiming to generate images from audio, including 9 categories from SoundNet~\citep{soundnet} and 5 categories from the VEGAS~\citep{vegas} datasets, respectively. 
Building on these foundations, Sound2Scene~\citep{sung2023sound} has made substantial progress in addressing sound-to-image generation problem with enhanced audio-visual latent alignment. They show the model's capabilities in handling more categories, surpassing prior art by generating more realistic images, and demonstrating control over the generation process. 
Subsequent works~\citep{audiotoken, qin2023gluegen} have utilized the generative capabilities of the Latent Diffusion Model~\citep{stable} for the sound-to-image generation task.

This work extends Sound2Scene~\citep{sung2023sound} by providing further analysis on the multimodal gap of aligned audio-visual latent space and conducting additional experiments to demonstrate the proposed method's generalizability in terms of architectural design and training dataset type.

\paragraph{Audio-driven image manipulation} 
Recent work has demonstrated the ability to \emph{edit} a reference image using input sound, rather than directly \emph{generating} a new one. Lee~\emph{et al}\onedot~\citep{soundguide} adapt a text-based image manipulation model~\citep{styleclip}, extending its embedding space to include audio-visual modalities alongside text. 
Similarly, Li~\emph{et al}\onedot~\citep{li2022learning} employ conditional generative adversarial networks (GANs)~\citep{gan} to manipulate the visual style of an image to match specific input sounds, showing that these modifications can be controlled by varying the sound's volume or by blending multiple sounds. 
Unlike these previous works, our approach focuses on \emph{generating} images based on sound inputs, with the \emph{editing} capability emerging as a byproduct of our method.

\begin{figure*}[tp]
    \centering
    \includegraphics[width=\linewidth]{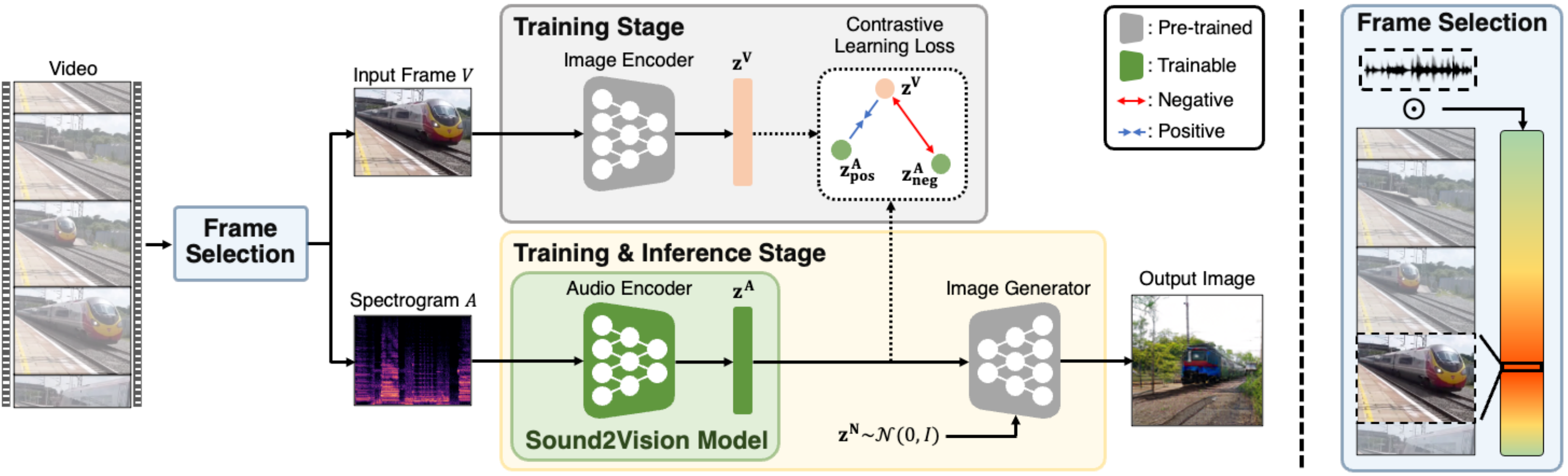}
    \caption{{\bf Sound2Vision framework.} First, the frame selection method selects the highly correlated frame-audio segment from a video for training. Then, we train Sound2Vision to produce an audio feature that aligns with the visual feature extracted from the pre-trained image encoder. In the inference stage, the extracted audio feature from input audio is fed to the image generator to produce an image.}
    \label{fig:pipeline}
\end{figure*}

\paragraph{Cross-modal generation}
Translating one modality into another, \emph{i.e.}, cross-modal generation, remains an interesting yet open research challenge. 
Various tasks across diverse modalities have been explored, including text-to-image/video~\citep{dalle,dalle2,styleclip,cogview, text2video1, text2video2, text2video3, stable}, touch-to-image~\citep{yang2023generating,yang2024binding}, speech-to-motion~\citep{speech2gesture, sung2024multitalk,sung2024laughtalk}, and image/audio-to-caption~\citep{sungbin, clipcap, flamingo}, among others. 
To bridge the gap between heterogeneous modalities in cross-modal generations, several studies~\citep{speech2face,dreamfusion} have leveraged existing pre-trained models or extended the pre-trained CLIP~\citep{clip} embedding space, which is anchored in the text-visual modality, to meet their specific requirements~\citep{styleclip,soundguide,dalle2, clipactor}. 
In this context, our work aims to generate images from sound by leveraging only freely acquired audio-visual signals from videos.

\paragraph{Audio-visual learning}
The natural co-occurrence of audio and visual cues is frequently used as a self-supervision signal to learn the associations between the two modalities, thereby enhancing representation learning. 
These learned representations are then utilized across various applications including cross-modal retrieval~\citep{objects,owens2018learning,senocak2024aligning}, video recognition~\citep{chen2021distilling,agreement}, and sound source localization~\citep{less,senocak2018learning,localize19,hardway,park2022marginnce,park2024can,senocak2023sound}. 
One approach to constructing an audio-visual embedding space involves jointly training separate neural networks for each modality to determine whether corresponding frames and audio match~\citep{look,owens18}. 
Recent efforts have incorporated clustering~\citep{hu2019deep,alwassel2020self} or contrastive learning~\citep{morgado2021robust,agreement,vatt} to learn this joint audio-visual embedding space more effectively.
While many of the aforementioned methods develop audio-visual representations jointly from scratch, another stream of research builds a joint embedding space by leveraging pre-existing expert models. 
This knowledge transfer may occur from audio to visual representations~\citep{owens2016ambient}, from visual to audio representations~\citep{soundnet,gan2019self}, or be distilled from both audio-visual representations to video-specific representations~\citep{chen2021distilling}.
Our research aligns with this latter approach, where the visual expert model is initially trained to construct an anchored space. This visual expert model is used to distill detailed visual information from extensive Internet videos into the audio modality.

\section{Method}\label{method}

In this section, we provide a concise overview of our proposed method for the sound-to-image generation task, followed by details on the training method. We then discuss the network architecture of our proposed model. Finally, we introduce a novel method for constructing highly correlated audio-visual data pairs from in-the-wild videos for this task.

\paragraph{Overview}
Our objective is to develop a method for translating input sounds into visual scenes. 
Existing methods~\citep{towards_a2s,chen2017,s2i,cmcgan} typically train generative models to synthesize images directly from raw audio or processed audio features without enforcing \textit{audio-visual alignment}. However, these approaches face significant challenges in producing high-quality, recognizable images due to the substantial gap between modalities and the inherent complexity of visual scenes.

To tackle these issues, we approach this task by breaking it down into sub-problems. The overall framework of our proposed model, Sound2Vision, is shown in Fig.~\ref{fig:pipeline}. This framework consists of three main components: an audio encoder, an image encoder, and an image generator. Initially, we separately pre-train a strong image encoder and an image generator conditioned by the embeddings of the image encoder with a large-scale image-only dataset. 
Given the inherent correlation between co-ocurring audio-visual signals, we leverage this natural alignment to infuse the audio features with rich visual information extracted by the image encoder. 
This process enables the construction of an aligned audio-visual latent space, trained through self-supervised learning on in-the-wild videos. This alignment gives cross-modal transferability~\citep{zhang2023diagnosing}. Consequently, the enriched audio features from this aligned latent space are fed into the image generator, allowing it to produce visual scenes that accurately reflect the given sounds.

\subsection{Learning to generate images from sound} \label{ssec:training}
Given the audio and image data pairs $\mathcal{D}=\{V_i, A_i\}^N_{i=1}$, where $V_i$ represents a video frame and $A_i$ its corresponding audio, the learning objective of Sound2Vision is to train the audio encoder to extract audio features $\mathbf{z^A}$ that align with anchored visual features $\mathbf{z^V}$. 
Specifically, the unlabeled data pairs $\mathcal{D}$ are fed into the audio encoder $f_{A}(\cdot)$ and the image encoder $f_{V}(\cdot)$ to respectively extract audio features $\mathbf{z^A}{=}f_{A}(A)$ and visual features $\mathbf{z^V}{=}f_{V}(V)$, where $\mathbf{z^V},\mathbf{z^A}\in R^{2048}$. 
Since the image encoder $f_{V}(\cdot)$ is well pre-trained on an image-only dataset, the visual feature $\mathbf{z^V}$ from the image encoder serves as the self-supervision signal for the audio encoder.
This facilitates the alignment of the audio feature with the visual feature through feature-based knowledge distillation~\citep{hinton, kdsurvey}.
These feature alignments across modalities
construct the shared audio-visual embedding space on which
the image generator G(·) is separately trained compatibly.

To align features from heterogeneous modalities, a metric learning approach is commonly employed.
This approach assumes that features are aligned if they are close to each other under some distance metric. 
One straightforward method is to minimize the $L_2$ distance $\norm{\mathbf{z^V}-\mathbf{z^A}}_2$ to align the features of $\mathbf{z^A}$ and $\mathbf{z^V}$.
However, we discover that relying solely on $L_2$ loss can capture the relationship between two different modalities within each pair without considering unpaired samples. 
This limitation may lead to constructing an aligned space that is not sufficiently rich, resulting in the generation of lower-quality images.
Therefore, we utilize InfoNCE~\citep{infonce}, a type of contrastive learning that has proven effective in learning audio-visual representation~\citep{afouras2020AVObjects,hardway,arda2022, chen2021distilling, wav2clip}:
\begin{equation}
\texttt{InfoNCE}(\ba_j, \{\bb\}_{k=1}^N) = -\log{\tfrac{\exp(-d(\ba_j,\bb_j))}{\sum^N_{k=1}\exp(-d(\ba_j,\bb_k))}},
\end{equation}
where $\ba$ and $\bb$ denotes arbitrary features with the same dimension, and $d(\ba,\bb)=\norm{\ba-\bb}_2$. 
By applying this loss, we aim to maximize feature similarity between an image and its corresponding audio segment (positive) while minimizing similarity with randomly selected, unrelated audio samples (negatives).
More specifically, for the $j$-th visual and audio feature pair, we define our audio feature-centric loss
as $L_{j}^{A} = \texttt{InfoNCE}(\mathbf{\hat{z}^A_j}, \{\mathbf{\hat{z}^V}\})$,
where $\mathbf{\hat{z}^A}$ and $\mathbf{\hat{z}^V}$ represent the unit-norm features.
To make a symmetric learning objective, we also compute the visual feature-centric loss term as
$L_{j}^{V} = \texttt{InfoNCE}(\mathbf{\hat{z}^V_j}, \{\mathbf{\hat{z}^A}\})$.
The overall learning objective is to minimize the sum of loss terms across all audio-visual pairs in mini-batch $B$:
\begin{equation}\label{loss1}
    L_{total} = \textstyle\frac{1}{2B}\sum\nolimits_{j=1}^B\left(L_j^A+L_j^V\right).
\end{equation}

After training the audio encoder using \Eref{loss1}, our model successfully learns to extract the audio features that are visually enriched and aligned with corresponding visual features. 
Thus, in the inference stage, we can directly feed the learned audio feature $\mathbf{z^A}$ along with a noise vector $\mathbf{z^N}\sim \mathcal{N}(0,I)$ to the frozen image generator, $G(\mathbf{z^N}, \mathbf{z^A})$, to generate a visual scene from the input sound. This is possible because our training objective enables cross-modal transferability, allowing audio features to replace image features.

\subsection{Architecture details}\label{sec:architecure}
All the following modules are trained separately according to the proposed steps. The modules presented here are the default design choices. However, our generic pipeline is versatile and can be applied to different architectural choices. This will be discussed in Sec.~\ref{sec:dif}.

\paragraph{Image encoder $f_V(\cdot)$}
We use ResNet-50~\citep{he2016deep} for the image encoder. To effectively handle a wide range of visual contents, we train the image encoder in a self-supervised manner~\citep{swav} using ImageNet~\citep{imagenet} without relying on labels.

\paragraph{Image generator $G(\cdot)$}
We utilize BigGAN~\citep{biggan} architecture to generate images with diverse visual scene contents.
We adapt the input structure based on modifications from ICGAN~\citep{icgan} to make BigGAN as a conditional generator.
This setup allows us to train the generator to produce photo-realistic images at a resolution of $128\times128$ using the conditional visual features $\mathbf{z^V}$ obtained from the image encoder.
The generator is trained on ImageNet in a self-supervised manner without labels, while the image encoder is pre-trained and remains fixed.

\paragraph{Audio encoder $f_A(\cdot)$}
We use ResNet-18, which takes the audio spectrogram as input.
Following the final convolutional layer, adaptive average pooling aggregates the temporal-frequency information into a single vector. 
This pooled feature is then fed into a linear layer to produce the audio feature $\mathbf{z^A}$. The audio encoder is trained using either the VGGSound~\citep{vggsound} or VEGAS~\citep{vegas} datasets, applying the loss defined in \Eref{loss1}, according to each target benchmark.

\begin{figure}[tp]
    \centering
    \includegraphics[width=\linewidth]{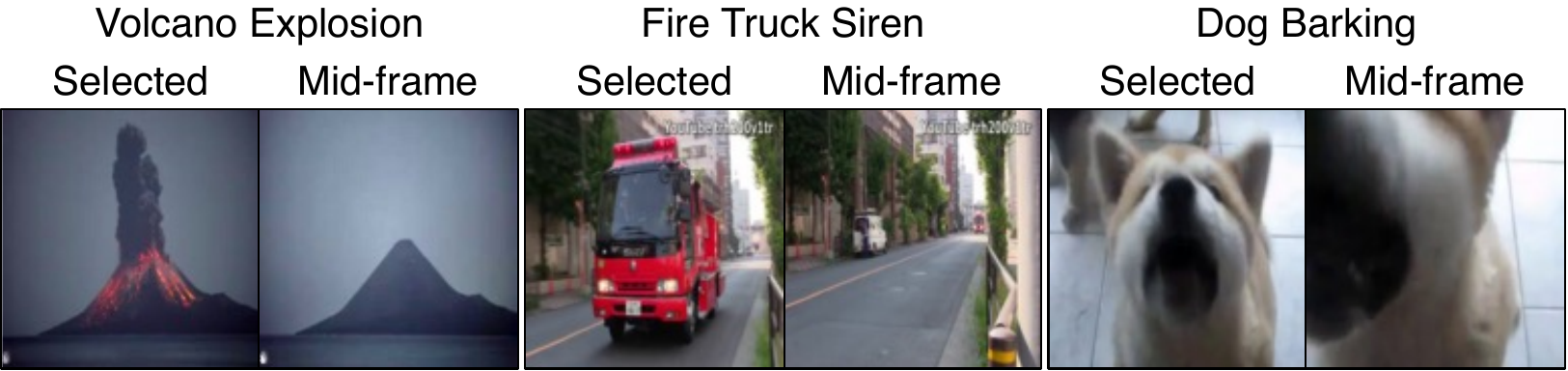}
    \caption{\textbf{Examples of comparison between the selected \texttt{top-1} frame versus mid-frame in the video.}}
    \label{fig:pair}
\end{figure}

\subsection{Audio-Visual pair selection module} \label{sssec:data}
Learning the relationship between images and sounds accurately requires highly correlated data pairs from both modalities. Identifying the most informative frame or segment in a video for audio-visual correspondence is not a straightforward task.
One simple way to collect audio-visual pairs $\mathcal{D}$ is to extract the mid-frame of a video along with the corresponding audio segment~\citep{soundguide,hardway}. 
However, the mid-frame does not always guarantee the presence of informative audio-visual signals~\citep{less}, as in Fig.~\ref{fig:pair}.
To this end, we leverage a pre-trained sound source localization model~\citep{less} to extract highly correlated audio and visual pairs. 
The backbone networks of~\citep{less} enable us to have fine-grained audio-visual features of temporal time steps, $\mathbf{q^A}$ and $\mathbf{q^V}$, respectively. 
The correlation scores are computed for each time step as $\mathbb{C}_{av}[t]=\mathbf{q^V_t}\boldsymbol{\cdot} \mathbf{q^A_t}$, then sorted by \texttt{top-k}($\mathbb{C}_{av}[t]$). 
Using this method of selecting correlated pairs, we annotate the \texttt{top-1} moment frames for each video in the training splits and use them for training. Figure~\ref{fig:pair} demonstrates a comparison between selected frames and mid-frames. Although selected automatically, they consistently contain distinct or salient objects that accurately correspond to the audio.

\section{Experiment}\label{experiment}
We evaluate our proposed Sound2Vision through a series of qualitative and quantitative assessments.
First, we provide qualitative analysis of the images generated from various categories of sound. Then, we quantitatively assess the quality, diversity, and correspondence between the audio and the generated images. 
Importantly, we do not use any class labels during training or inference.

\subsection{Experiment setup}\label{sec:setup}
\paragraph{Datasets}
We utilize the VGGSound~\citep{vggsound} and VEGAS~\citep{vegas} datasets for training and testing our method. 
VGGSound contains approximately 200K videos, from which we chose 50 classes and follow the provided training and testing splits. 
VEGAS, on the other hand, includes about 2.8K videos with 10 classes. 
To maintain data balance, we use 800 videos for training and 50 videos for testing from each class. 
We use the test splits from both datasets for subsequent qualitative and quantitative evaluations.

\paragraph{Evaluation metrics}
We use both objective and subjective metrics to evaluate our method.

\begin{itemize}
    \item \textbf{CLIP}~\citep{clip} \textbf{retrieval} : Inspired by the CLIP R-Precision metric~\citep{clipr}, we evaluate the generated images by conducting an image-to-text retrieval test, measuring recall at $K$ ($R@K$). 
    We input the generated images along with audio category names (text) into CLIP, then measure the similarity between the image and text features to rank the text descriptions for each query image.    
    \item \textbf{Fr\'{e}chet Inception Distance (FID)}~\citep{fid} \textbf{and Inception Score (IS)}~\citep{is} : We use FID
     to measure the Fr\'{e}chet distance between features obtained from real and synthesized images using a pre-trained Inception-V3 model~\citep{inceptionv3}. 
    This same model is also used IS, computing the KL-divergence between the conditional and marginal class distributions.    
    \item \textbf{Human evaluations} : We recruit 70 participants to analyze the performance of our method from a human perception perspective. 
    Participants are asked to compare our model with an image-conditioned generation model~\citep{icgan} and to assess whether the images generated by our model accurately correspond to the input sounds. Further details are available in Sec.~\ref{ssec:quan}. 
\end{itemize}

\paragraph{Implementation details}
The audio encoder takes a $1004 \times 257$-dimensional log-spectrogram, which is converted from 10 seconds of audio, to extract audio features. Simultaneously, the video frame is resized to $224 \times 224$ and fed into the image encoder to extract visual features.
We train our model on a single GeForce RTX 3090 for 50 epochs with early stopping. The Adam optimizer is used, with a batch size set at 64, a learning rate of $10^{-3}$, and a weight decay of $10^{-5}$.


\begin{figure}[tp]
    \centering
    \includegraphics[width=1\linewidth]{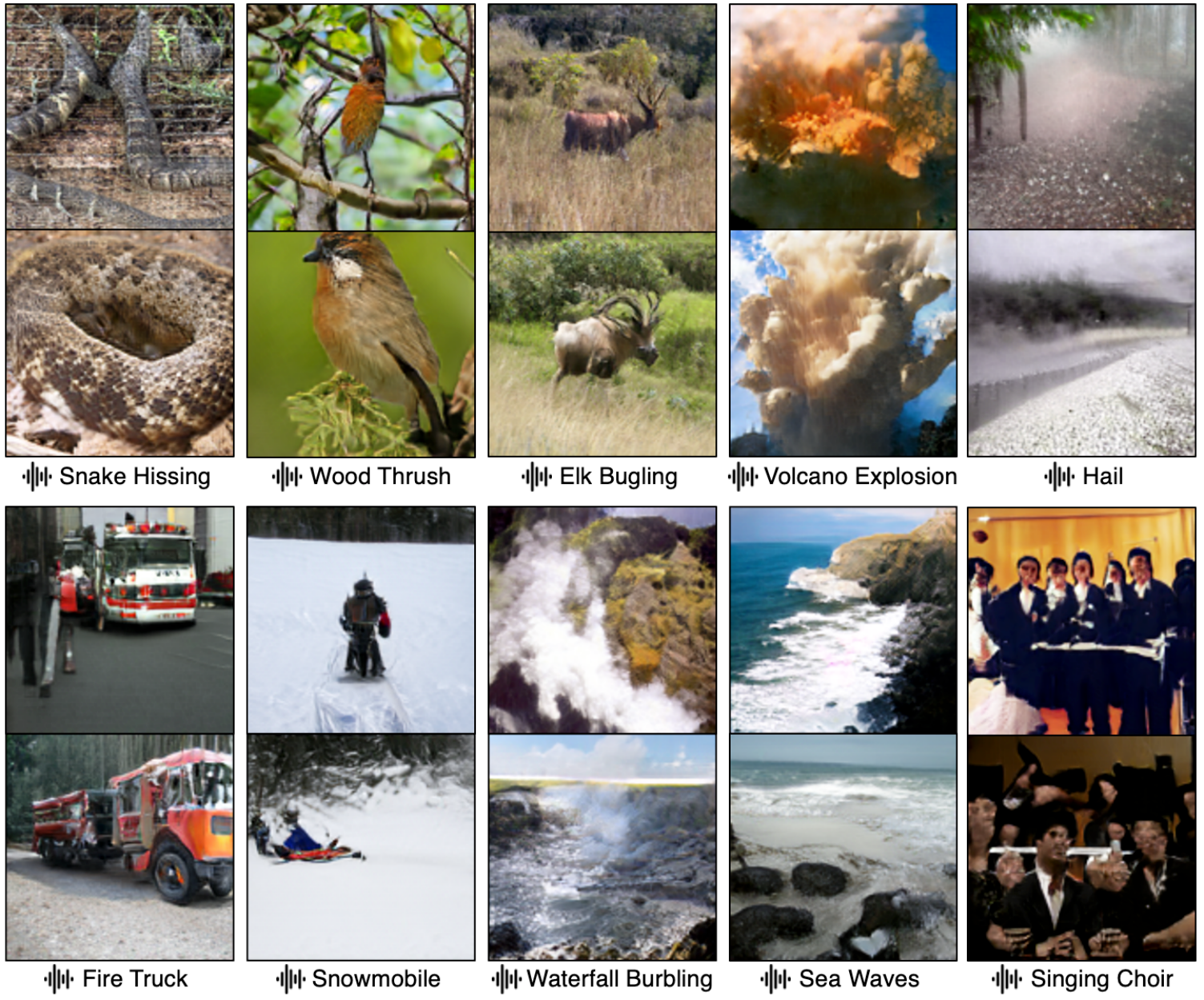}
    \caption{\textbf{Qualitative results by feeding single waveform from VGGSound test set.} Sound2Vision generates diverse images in a wide variety of categories from generic sounds as input.}
    
    \label{fig:single}
\end{figure}

\begin{figure}[tp]
    \centering
    \includegraphics[width=0.9\linewidth]{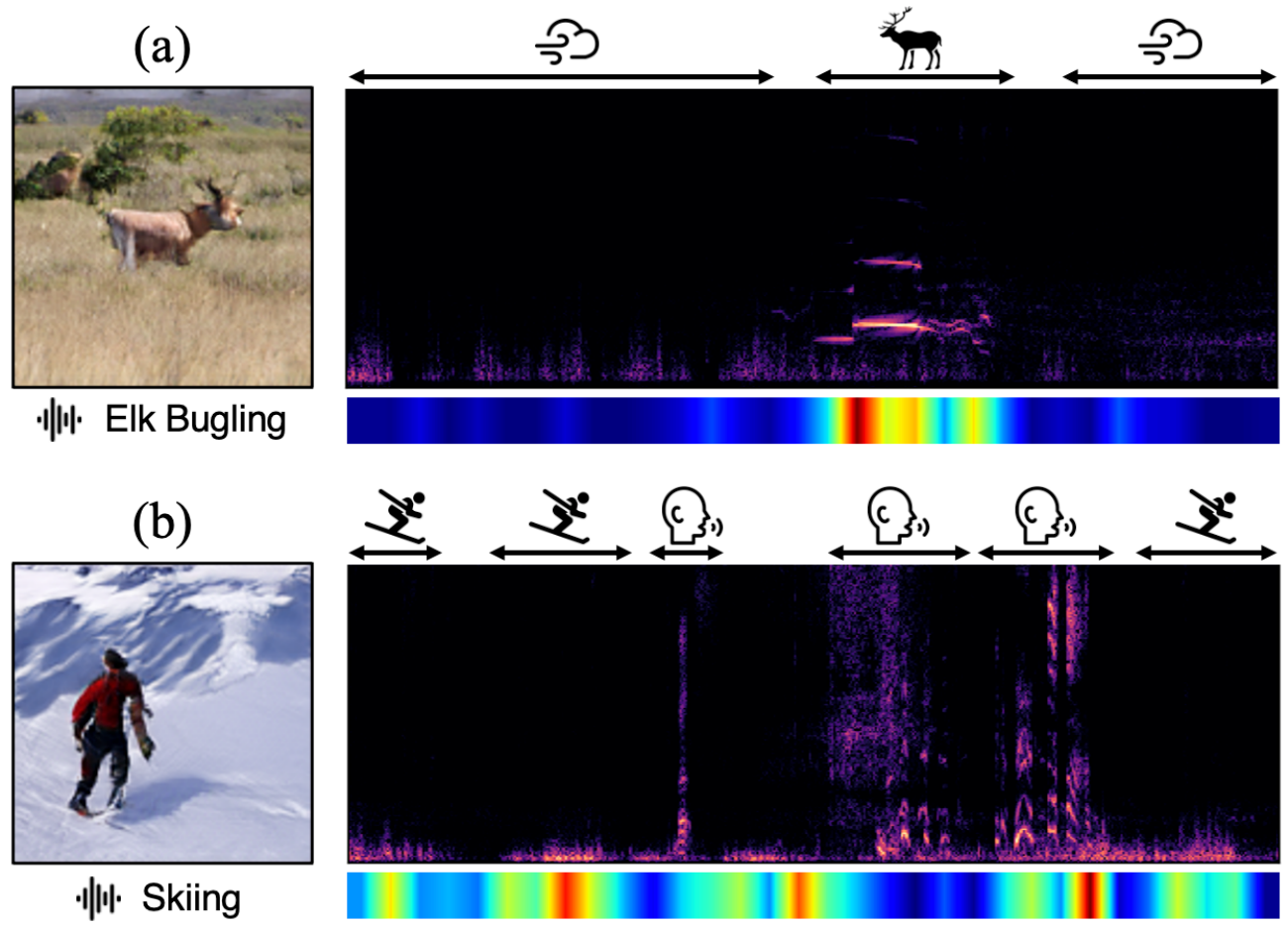}
    \caption{\textbf{Grad-CAM~\citep{gradcam} visualization for the highlighted moment in the spectrograms.}
In the heatmap, regions most highlighted during image generation are colored red, transitioning to blue in less highlighted areas.
\windEmoji, \elkEmoji, \skiEmoji, and \humanEmoji \,\,denote wind blowing, elk bugling, skiing, and human talking sounds, respectively.}
    
    \label{figure:gradcam}
\end{figure}

\subsection{Qualitative analysis}
\paragraph{Image generation from sound}
Sound2Vision generates visually plausible images from a single input waveform, as shown in Fig.\ref{fig:teaser} and \ref{fig:single}.
Unlike previous work~\citep{s2i,towards_a2s}, it is not limited to a small number of categories but instead handles a diverse range, including animals, vehicles, sceneries, \emph{etc}\onedot
We highlight that our model can even distinguish subtle differences in similar sound categories, such as ``Waterfall Burbling'' and ``Sea Waves'' sounds, and produces accurate and distinct images.

\paragraph{Audio event visualization}
After presenting the results of Sound2Vision in generating a wide variety of images, we analyze which parts of the audio the model uses to generate images and whether it is attending to the true ``semantic'' context in the audio.
With the trained Sound2Vision, we visualize a coarse localization map on the spectrogram, indicating areas the model highlights for image generation. 
Using the same technique of Gradient-weighted Class Activation Mapping (Grad-CAM)~\citep{gradcam}, we compute the gradient of the generated image with respect to the feature map activation of the audio encoder's last convolutional layer to produce the coarse localization map (Fig.~\ref{figure:gradcam}). As shown, each highlighted moment is visualized in a heatmap, with the most emphasized regions transitioning from blue to red.
For example, (a) shows that the model focuses on a certain region of the sound that is dominant throughout the duration, \emph{e.g.}, elk bulging, for generating an image.
On the other hand, (b) demonstrates that the model focuses on both regions of skiing sound (wind blowing and footsteps on snow) and human sound, to produce a composite image.


\subsection{Quantitative analysis}
\label{ssec:quan}

\begin{table*}[t]
\footnotesize
\centering
\caption{\textbf{Quantitative evaluations.} We compare our method with different baselines (different settings for the encoder and the generator) on CLIP retrieval~(R@k), FID, and IS in (a). 
        For user study, we first compare our method with ICGAN by measuring recall probability between generated images of ICGAN and our method from the same audio-visual pair. Second, we validate the correspondence our method’s output for the given audio. Results are in (b) respectively.
        $\emph{Abbr.}$ $V$: image encoder, $A$: audio encoder, $G$: image generator, $R$: retrieval system.}
        
    \resizebox{0.68\linewidth}{!}{
    \begin{tabular}{ll @{\quad}cc@{\quad}cccc@{\quad}cc}
    \toprule
    &\multirow{2}{*}{Method}& \multirow{2}{*}{\begin{tabular}[c]{@{}c@{}}Encoder\\ ($V$/$A$)\end{tabular}}&  \multirow{2}{*}{\begin{tabular}[c]{@{}c@{}}Generator\\ ($G$/$R$)\end{tabular}}&\multicolumn{4}{c}{VGGSound (50 classes)}&\multicolumn{2}{c}{VEGAS}\\
    \cmidrule(r{4mm}){5-8} \cmidrule{9-10}
     &&&& R@1 & R@5 & FID ($\downarrow$) & IS ($\uparrow$) & R@1& R@5 \\
    \cmidrule{1-10}
    (A)&ICGAN~\citep{icgan} & $V$ & $G$ & 30.06 & 62.59 & \textbf{16.11} & 12.61 &46.60 & 82.48 \\
    (B)&Ours & $A$ & $G$ & \textbf{40.71}  & \textbf{77.36} & 17.97 & \textbf{19.46} & \textbf{57.44} & \textbf{84.08} \\
    \cmidrule{1-10}
    (C)&Retrieval & $A$ & $R$ & 51.28 & 80.37 & - & - & 67.20 & 85.00 \\
    (D)&Upper bound &-&-& 57.82 & 85.79 & - & -&73.60 & 88.2\\
    \bottomrule
    \end{tabular}
    }
    \blank{0.2cm}
    \resizebox{0.28\linewidth}{!}{
    \begin{tabular}{c}
    \includegraphics[width=1\linewidth]{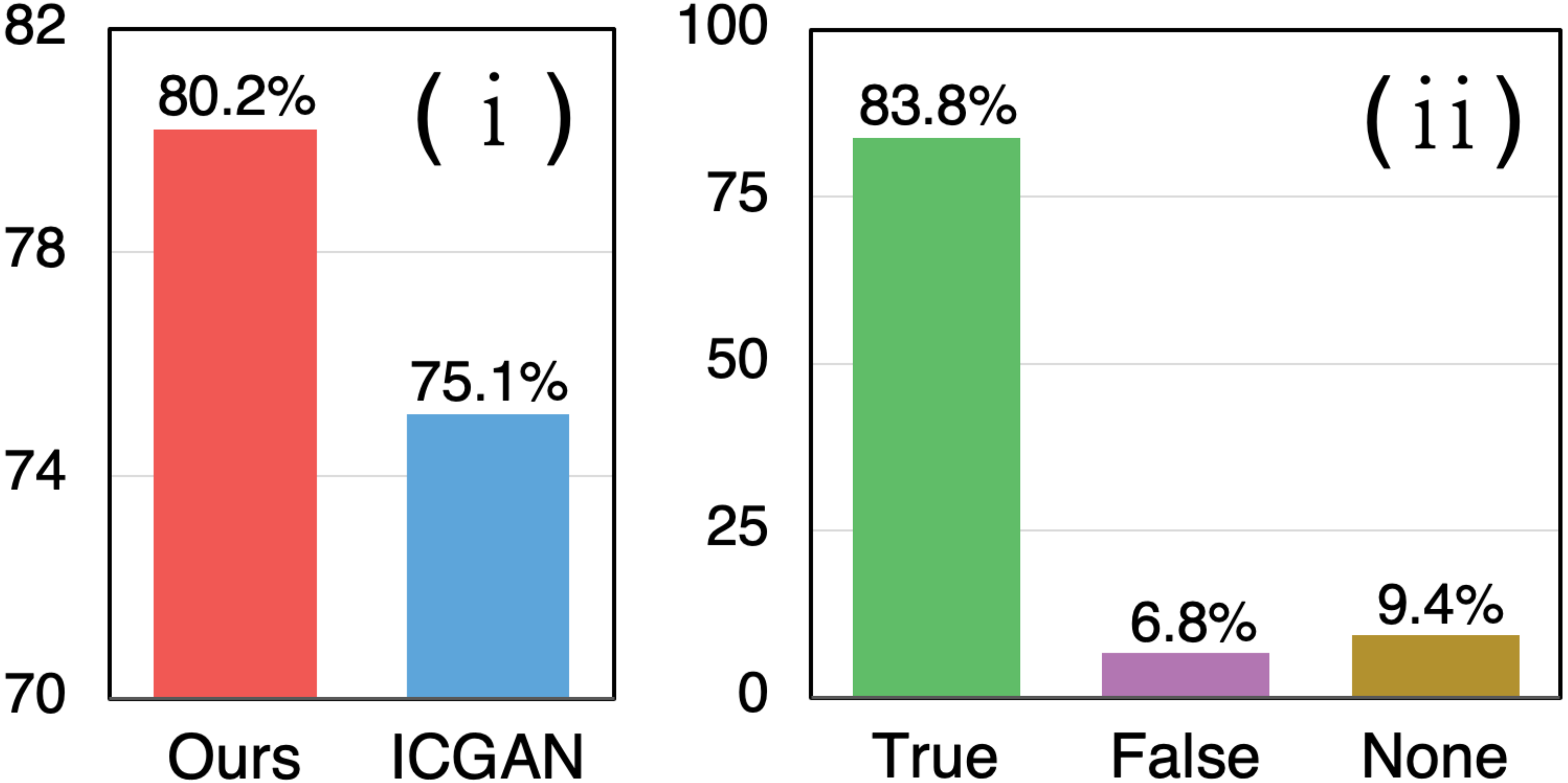}
    \end{tabular}}\\
    
    (a) Comparison to baselines \blank{6.3cm} (b) User study\blank{-2.4cm}
     \vspace{-3mm}
\label{tab:baseline}
\end{table*}

\paragraph{Comparison with baselines}
 We conduct a series of experiments to validate our proposed method on VGGSound and VEGAS, compared with several closely related baselines, as detailed in ~Table~\ref{tab:baseline}~(a). 
First, our model (B) is compared with an image-to-image generation model identical to ICGAN\citep{icgan} (A). 
Although (B) shares the same image generator as (A), it utilizes a different encoder type and input modality. 
Results show that (B) performs favorably against (A) across all metrics. 
We attribute this to the noisy nature of video datasets, which may disturb (A) from extracting informative visual features for image generation, whereas audio input proves to be more robust to these disturbances, thus resulting in more plausible images. 
Additionally, we evaluate our model (B) against a retrieval system (C) that serves as a strong baseline. 
The retrieval system uses the same audio encoder as (B), while the image generator $G$ is replaced with the same memory-sized database of images from the training data.
This system finds the closest image from the database, given the input audio feature.
Compared to (D)—an upper bound where video frames are directly used—
(C) shows a significantly smaller performance gap with (D) compared to the gap between (B) and (D), validating the 
effectiveness of our audio encoder in mapping audio to the shared embedding space. While (C) surpasses (B) in $R@1$ for both datasets, (B) performs comparably with (C) in $R@5$, demonstrating that our method approaches the performance of this strong baseline.

\paragraph{User study}
The user study results are summarized in Table~\ref{tab:baseline}~(b) across two experiments: (\romannumeral1) a comparison with ICGAN, and (\romannumeral2) an evaluation of the plausibility of image generation from given audio. 
Each experiment consists of 20 questions where participants are presented with audio and multiple images. 
In (\romannumeral1), participants select images they believe best represent the audio, with two images generated by both our model and ICGAN, and the remaining images randomly chosen from either method. 
We measure preference by comparing the recall probability of ICGAN with our model. 
In (\romannumeral2), all four provided images are generated by our model, but only one corresponds to the given audio. Participants then select the image they find most related to input audio. 
In (\romannumeral1), our model is preferred. 
Additionally, (\romannumeral2) demonstrates that our method achieves a precision rate of 83.8\%, supporting that our model generates images highly correlated with the provided audio.

\begin{table}[t]
\vspace{1.5mm}
\footnotesize
\centering
\caption{\textbf{Ablation studies of our proposed method.} We compare different configurations of our method by changing the loss functions, frame selection method (denoted as $F$), and duration of the audio. 
     }
    \resizebox{0.95\linewidth}{!}{
    \begin{tabular}{l@{\quad}l@{\quad}c@{\quad}c@{\quad}c@{\quad}c@{\quad}c@{\quad}c}
    \toprule
    \multirow{2}{*}{}&\multirow{2}{*}{Loss}&\multirow{2}{*}{$F$}
    &\multirow{2}{*}{Duration}
    &\multicolumn{4}{c}{VGGSound (50 classes)}\\
   
    \cmidrule{5-8}
     & &&& R@1 & R@5 & FID ($\downarrow$) & IS ($\uparrow$)\\
    \cmidrule{1-8}
    (A) &$L_2$ & \checkmark&10 sec.& 18.21 & 46.69 & 24.05 & 9.97\\
    (B) & $L_{nce}$& \checkmark&10 sec.& 31.63 & 66.04 & 27.05 & 12.92\\
    (C) & $L_{total}$& & 10 sec.&37.20 & 73.13 & 21.20 & 17.51\\
    \cmidrule{1-8}
    (D) &\multirow{2}{*}{$L_{total}$}&\multirow{2}{*}{\checkmark}&1 sec. &35.85& 72.02&19.05&17.87\\
    (E) &&&5 sec. &38.24&75.76&20.43&18.81\\
    \cmidrule{1-8}
    (F) & $L_{total}$& \checkmark& 10 sec.& \textbf{40.71} & \textbf{77.36} & \textbf{17.97} & \textbf{19.46}\\
    \bottomrule
    \end{tabular}
    }
     \vspace{-2mm}
\label{tab:ablation}
\end{table}

\paragraph{Ablation study}
The ablation studies conducted to assess various design choices are summarized in Table~\ref{tab:ablation}.
We evaluate the performance of using different distillation losses: a straightforward $L_2$ loss between the visual and audio features, and an InfoNCE loss~\citep{infonce} with cosine similarity distance measurement, $L_{nce}$, rather than using $L_2$ distance as in ~\Eref{loss1}. 
The results of experiments (A), (B), and (F) indicate that our chosen loss (F) leads the model to produce more diverse and improved image quality. 
Furthermore, the comparison between (C) and (F) demonstrates that our audio-visual pair selection method, as discussed in~Sec.~\ref{sssec:data}, contributes to performance improvements. 
Lastly, we examine the effect of audio duration by training models with 1, 5, and 10 seconds of audio while keeping other experimental settings the same. 
The results from (D), (E), and (F) show that longer audio durations consistently enhance performance, likely because longer ones may capture more comprehensive semantics, whereas shorter durations may miss critical details.


\subsection{Multimodal gap analysis}\label{sec:gap}
The key motivation of our work is to enable cross-modal transferability, allowing our audio embeddings to be used directly with visually pre-trained image generators. Our contrastive learning-based training objective facilitates cross-modal generation from sound to image, even though contrastive learning is known to result in a multimodal gap in the shared space~\citep{liang2022mind}. We analyze the properties of this multimodal gap that enable cross-modal generation and discuss how closing the gap could further enhance model performance.

\begin{figure}[tp]
    \centering
    \includegraphics[width=1\linewidth]{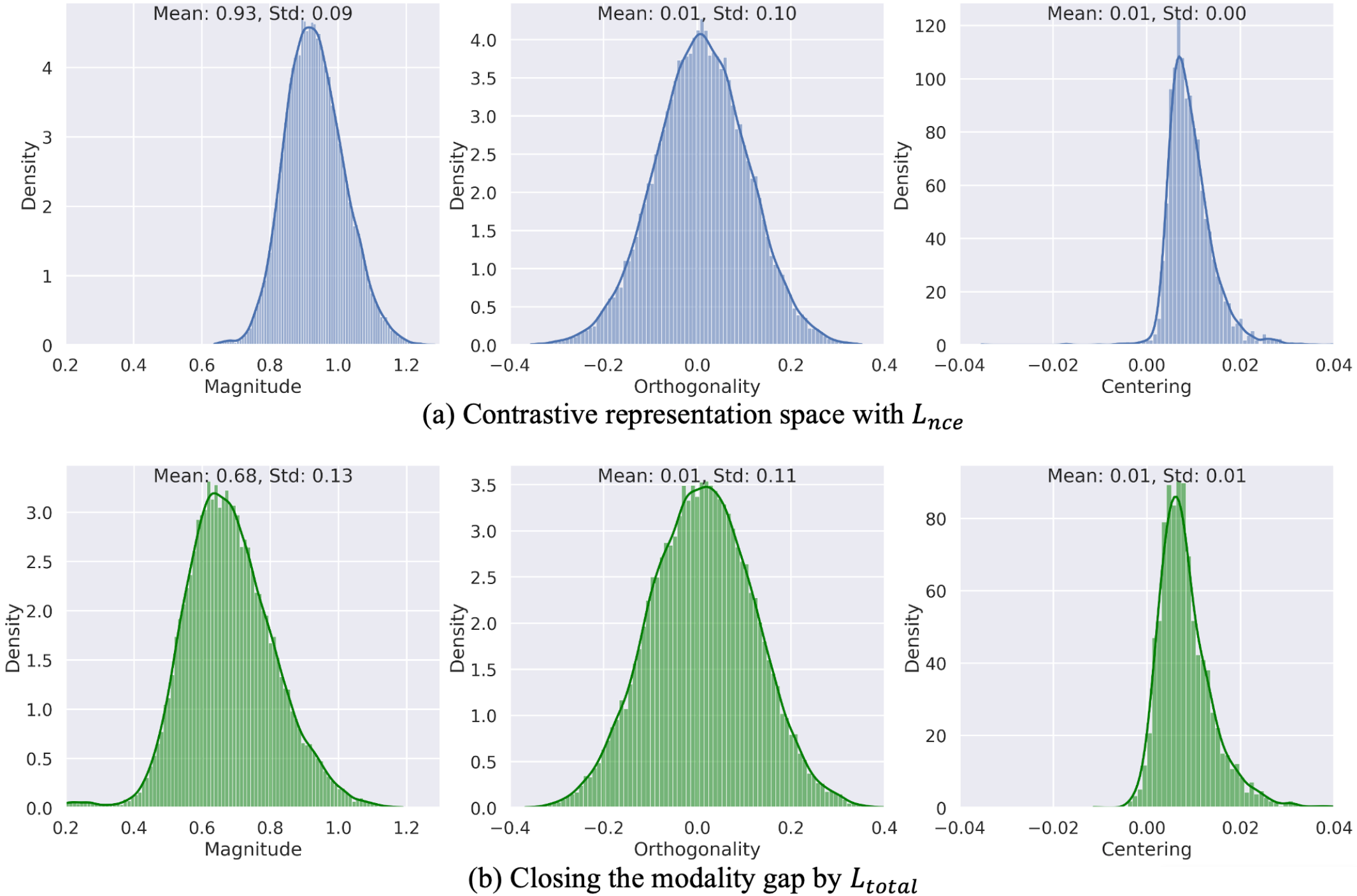}
    \caption{{\bf Geometry analysis of modality gap in learned audio-visual shared embedding space.} 
    We visualize the geometric properties of the modality gap in the contrastive representation space learned with (a) $L_{nce}$ and (b) $L_{total}$. 
    In both cases, the modality gap approximates a constant vector, as indicated by its magnitude. Furthermore, the modality gap is orthogonal to the span of features from both modalities, and the centers of features for each modality are zero vectors in the subspace orthogonal to the gap, as indicated by the orthogonality and centering distributions.}
    \label{fig:geometry}
\end{figure}

\paragraph{Cross-modal transferability}
As discussed in \citep{zhang2023diagnosing}, the cross-modal transferability is a phenomenon that allows the learned shared embedding space to make different modalities interchangeable for cross-modal tasks. According to Zhang~\emph{et al}\onedot, this intriguing phenomenon is enabled by the unique geometry of the modality gap:
\begin{itemize}
    \item The modality gap approximates a constant vector. This is verified by computing distributions over $\norm{\boldsymbol{g}}$ (magnitude), where $\boldsymbol{g}=\mathbf{z^V}-\mathbf{z^A}$ is the modality gap between the paired visual and audio features.
    \item The modality gap is orthogonal to the span of the features, and features have zero mean in the subspace orthogonal to the modality gap. We verify this by computing distributions over $\text{cos}(\mathbf{z^V}-\mathbb{E}_{\mathbf{z^V}}[\mathbf{z^V}],\mathbb{E}_{\boldsymbol{g}}[\boldsymbol{g}])$ (orthogonality) and $\mathbb{E}_{\mathbf{z^V}}[\mathbf{z^V}-(\mathbf{z^V})^T\boldsymbol{g'}\boldsymbol{g'}]_i$ (center), where $\boldsymbol{g'}=\mathbb{E}_{\boldsymbol{g}}[\boldsymbol{g}]/\norm{\mathbb{E}_{\boldsymbol{g}}[\boldsymbol{g}]}$ and $i$ denoting $i$-th dimension of each feature.
\end{itemize}
As demonstrated in Fig.~\ref{fig:geometry}~(a), we find that optimizing InfoNCE loss~\citep{infonce} with cosine similarity distance measurement, denoted as $L_{nce}$, results in a constant magnitude of the modality gap, while the orthogonality and centering values are near zero. 
This result supports the idea that the objective of our model preserves geometric properties that facilitate cross-modal transferability.

\paragraph{Closing the modality gap}
Although $L_{nce}$ objective has proven to be effective in learning aligned audio-visual features, we further explore how closing the multimodal gap affects sound-to-image generation performance. 
We compare different configurations of the loss function, from $L_{nce}$ to our final objective, $L_{total}$. 
Figure~\ref{fig:geometry}~(b) illustrates the geometry of the modality gap learned from $L_{total}$. 
The magnitude of the multimodal gap is significantly reduced compared to that using $L_{nce}$, while other properties, such as orthogonality and centering, remain close to zero.

\begin{figure}
    \centering
    \small
    \includegraphics[width=\linewidth]{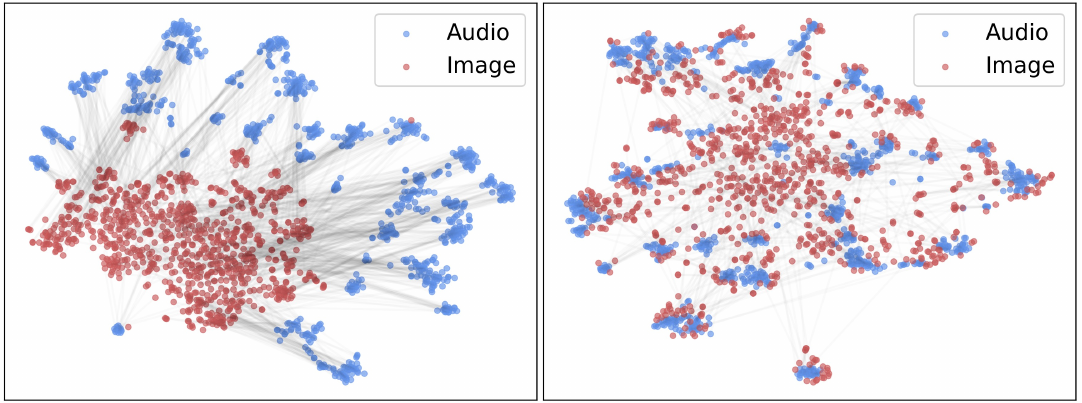}\\
    (a) $L_{nce}$ \quad\quad\quad\,\,\qquad\qquad\qquad(b) $L_{total}$\\
    \vspace{2mm}
    \resizebox{1 \linewidth}{!}{
    \begin{tabular}{lccccc}
    \toprule
    Method & R@1 & FID ($\downarrow$) & IS ($\uparrow$) & Magnitude ($\downarrow$)& Alignment ($\uparrow$)\\
    \cmidrule{1-6}
    
    (a) $L_{nce}$ &31.6 & 27.0 & 12.9 & 0.93 & 0.51\\
    (b) $L_{total}$ &\textbf{40.7} & \textbf{17.9} & \textbf{19.4} &\textbf{0.68} &  \textbf{0.71}\\
    \bottomrule
    \end{tabular}
    }
    \caption{{\bf Comparisons of audio-visual modality gap.} The aligned representation learned with two different loss functions, $L_{nce}$ and $L_{total}$, as illustrated in  Fig.~\ref{fig:geometry}, is visualized using a t-SNE~\citep{tsne}. We observe that reducing the modality gap between the audio and visual modalities, as indicated by the smaller multimodal gap in (b), leads to enhanced qualitative and quantitative performance, and improved alignment. \textit{Magnitude}~\citep{zhang2023diagnosing} refers to the size of the multimodal gap, while \textit{Alignment}~\citep{goel2022cyclip} indicates the degree of multimodal alignment.}
    
    \label{fig:gap}
\end{figure}

Building on this, we analyze the relationship between the modality gap and sound-to-image generation performance. 
Figure~\ref{fig:gap} shows the t-SNE~\citep{tsne} visualization of audio and visual features learned by different loss functions. 
While the features in Figure~\ref{fig:gap}~(a), learned with $L_{nce}$, exhibit a noticeable gap in the latent space, the features in Figure~\ref{fig:gap}~(b), learned with $L_{total}$, show less of a gap, with the two modalities overlapping. 
We observe significant improvements in overall metrics as we reduce the modality gap between audio-visual features, including quantitative metrics and the multimodal alignment measurement introduced in ~\citep{goel2022cyclip}.
This analysis suggests that learning to closely align audio features with visual features, particularly by reducing the gap between them, is essential for cross-modal generative tasks to produce diverse and visually plausible images.

\section{Controllability of the model}\label{sec:control}
Sound2Vision captures the natural correspondence between audio-visual signals through an aligned shared embedding space. Thus, we intuitively ask whether manipulations to the input can lead to corresponding changes in the generated images. 
Interestingly, we observe that our model supports controllable outputs through straightforward manipulations, either in the \emph{waveform space} or in the learned \emph{latent space}, even without a specific learning objective for such control. 
This opens up interesting experiments, which we explore in the following.

\subsection{Waveform manipulation for image generation}
\begin{figure}[tp]
    \centering
    \includegraphics[width=\linewidth]{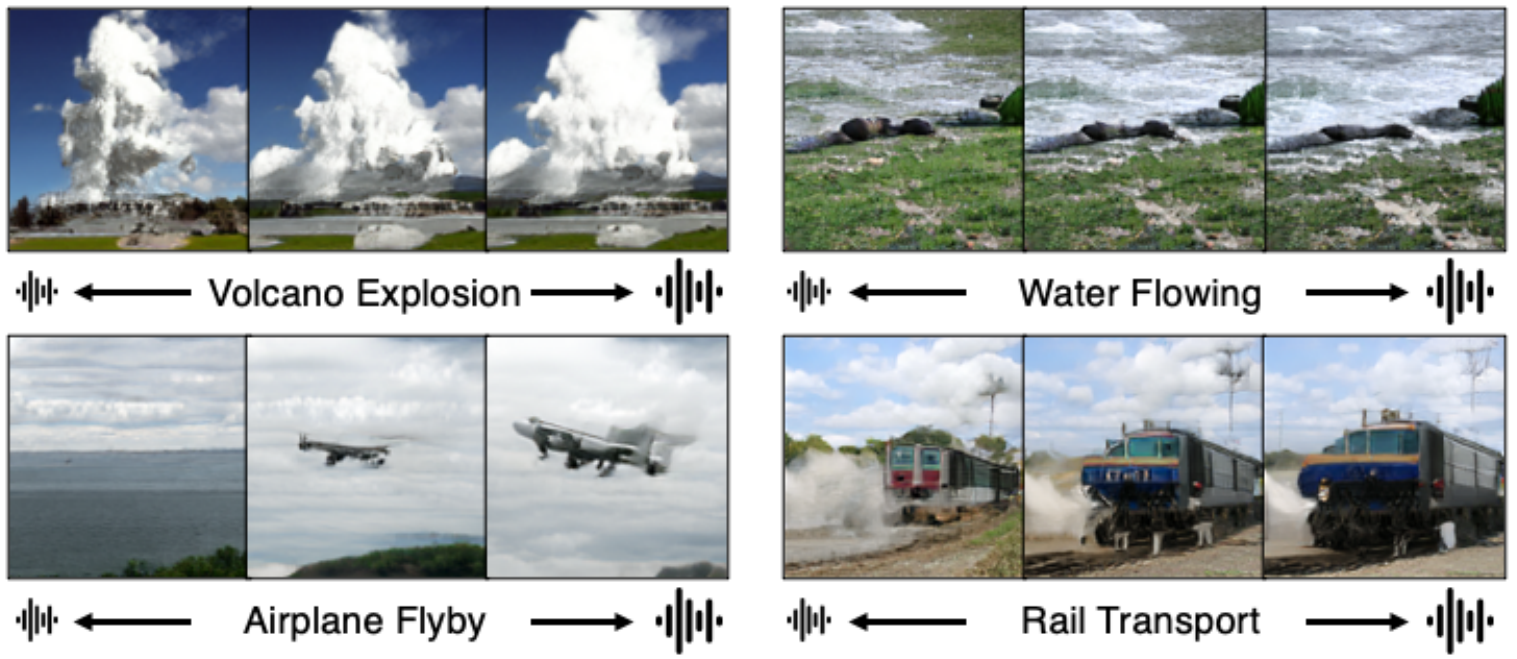}
    \caption{\textbf{Generated images by changing the volumes of the input audio in the \emph{waveform space}.} As the volume increases, the objects of the sound source become larger or more dynamic.}
    \label{fig:volume}
    
\end{figure}

\paragraph{Changing the volume}
Humans are capable of estimating the rough distance or size of an object based on the volume of its associated sound. 
To verify whether our model also has a similar ability to understand volume differences, we experiment by both reducing and increasing the volume of the reference audio. Each modified audio waveform input is fed into our model with the same noise vector. As shown in Fig.~\ref{fig:volume}, the objects in the generated images appear larger as the volume increases. 
Notably, in the ``Water Flowing'' example, volume adjustments result in images that depict varying strengths of water flow, while in the ``Rail Transport'' example, they illustrate a train appearing progressively closer in the scene.
These observations indicate our model's ability to not only recognize class-specific features but also understand the relationship between audio volume and visual changes. 
This suggests that visual supervision, rather than class labels, enables the model to capture such dynamic and meaningful audio-visual relationships.

\begin{figure}[tp]
    \centering
    \includegraphics[width=\linewidth]{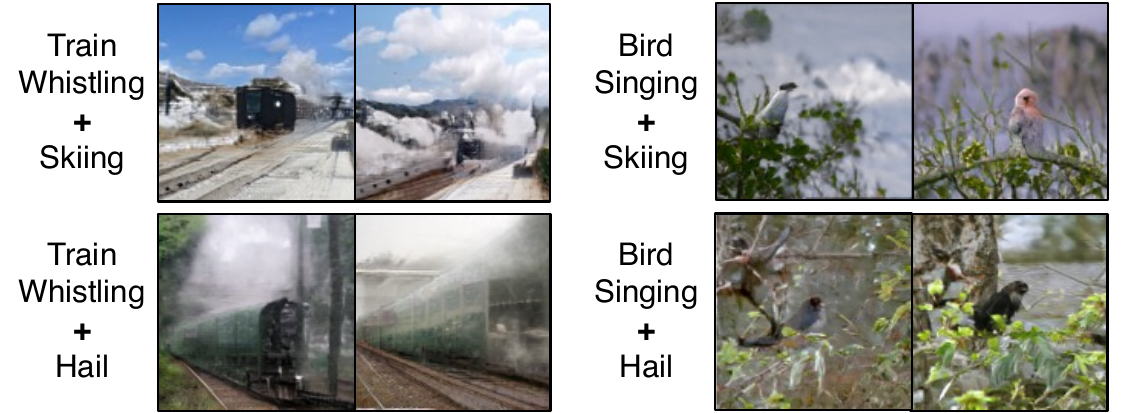}
    \caption{\textbf{Generated images by mixing two different audios in the \emph{waveform space}.}} 
    \label{fig:mix}
\end{figure}

\paragraph{Mixing waveforms}
We explore whether our model can reflect the presence of multiple sounds in a single generated image. To test this, we combine two distinct waveforms into one and input the mixed waveform into our model. As illustrated in~Fig.~\ref{fig:mix}, our model successfully generates images that capture the composite audio semantics. For example, mixing the ``Skiing'' sound with others results in images where elements such as a railroad or a bird emerge within a snowy scene. Similarly, when the ``Hail'' sound is mixed, the generated images show both a train and a bird appearing in a misty scene. The task of detecting multiple distinct sounds from a single combined audio input, known as audio source separation~\citep{gao2019co}, and visually representing them accurately within a single context is not trivial. 
Nevertheless, our findings indicate that our model is capable of achieving this to a certain extent.

\begin{figure}[tp]
    \centering
    \footnotesize
    \includegraphics[width=\linewidth]{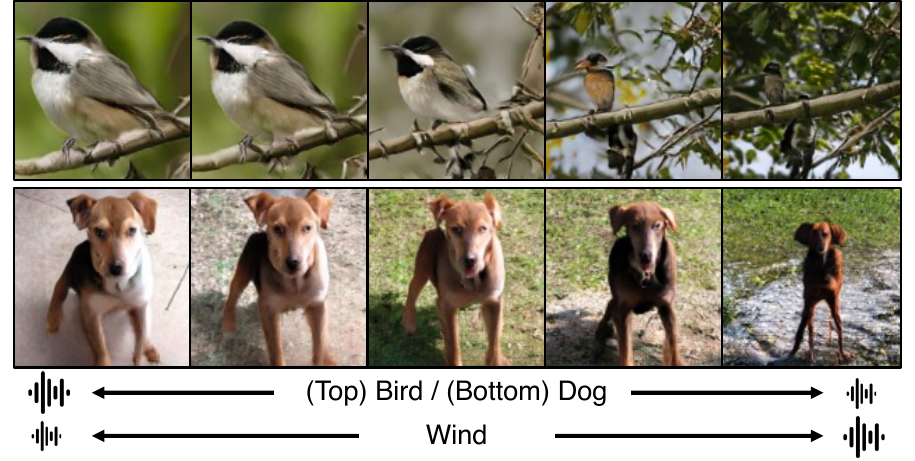}
    
    \caption{\textbf{Generated images by mixing multiple audios with volume changes in the \emph{waveform space}}. We observe that Sound2Vision mimics the camera movement by placing the object further as the wind sound gets larger.}
    \label{fig:mixvolume}
\end{figure}

\paragraph{Mixing waveforms and changing the volume}
In this experiment, we manipulate the input waveform by combining multiple waveforms and simultaneously adjusting the volume of each waveform.
In Fig.~\ref{fig:mixvolume}, we combine the ``Wind'' sound with either the ``Bird'' or ``Dog'' sounds, adjusting their volumes.
As the volume of the ``wind'' sound increases and the ``bird'' sound decreases, the bird appears smaller in the image, eventually getting covered by bushes.
Similarly, as the ``dog'' sound grows louder, the generated image transitions to a close-up of a dog indoors as the ``wind'' sound fades. However, when the ``wind'' sound becomes dominant again, the scene shifts to a wider shot of the dog outdoors.
These results demonstrate that our model can detect subtle changes in the audio and accurately reflect these variations in the generated images.

\subsection{Latent manipulation for image generation}
As we build an aligned audio-visual embedding space, our model can take both image and audio as inputs to generate images. 
We present two different approaches for audio-visual conditioned image generation, where both methods manipulate the features of the inputs in the latent space.

\begin{figure}[tp]
    \centering
    \small
    \includegraphics[width=\linewidth]{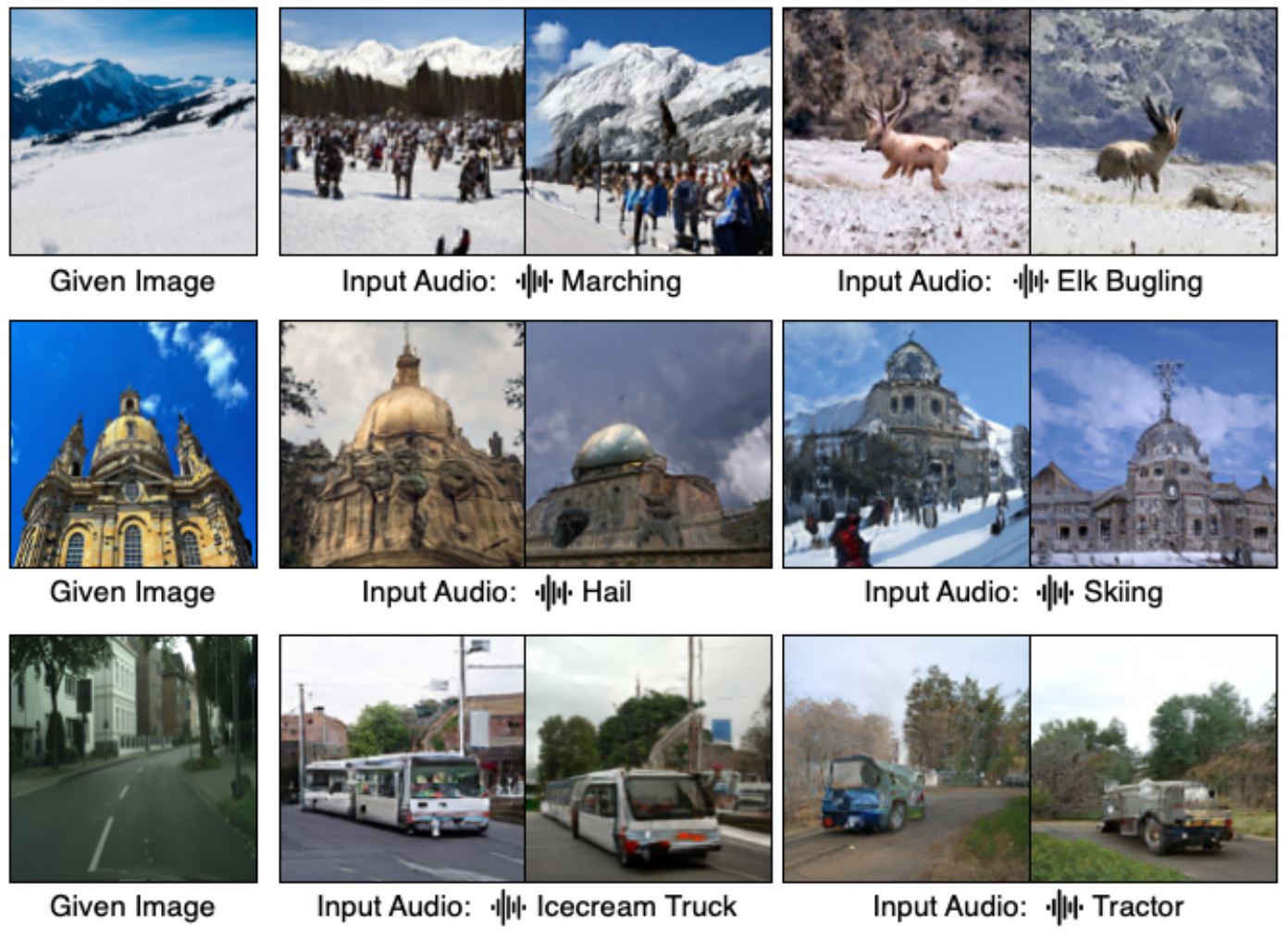}\\
    \blank{-1.4cm}(a) Inputs \blank{2.1cm} (b) Generated images
    \caption{\textbf{Generated images conditioned on image and audio.} We interpolate between a given visual feature and an audio feature in the \emph{latent space}. This interpolated feature is then fed to the image generator to get a novel image.}
    \label{fig:imgaud}
    
\end{figure}

\begin{figure}[t]
    \centering
    \footnotesize
    \resizebox{0.2\linewidth}{!}{
    \begin{tabular}{c}
    \includegraphics[width=0.4\linewidth]{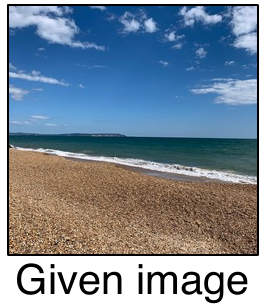}
    \end{tabular}}
    \resizebox{0.7\linewidth}{!}{
    \begin{tabular}{lc}
    \toprule
    Method&Target task\\
    \cmidrule{1-2}
    \citep{soundguide}&Sound-guided image manipulation\\
    Ours&Sound-to-image generation\\
    \bottomrule
    \end{tabular}
    }
    \vspace{1.5mm}
    \\
    \includegraphics[width=\linewidth]{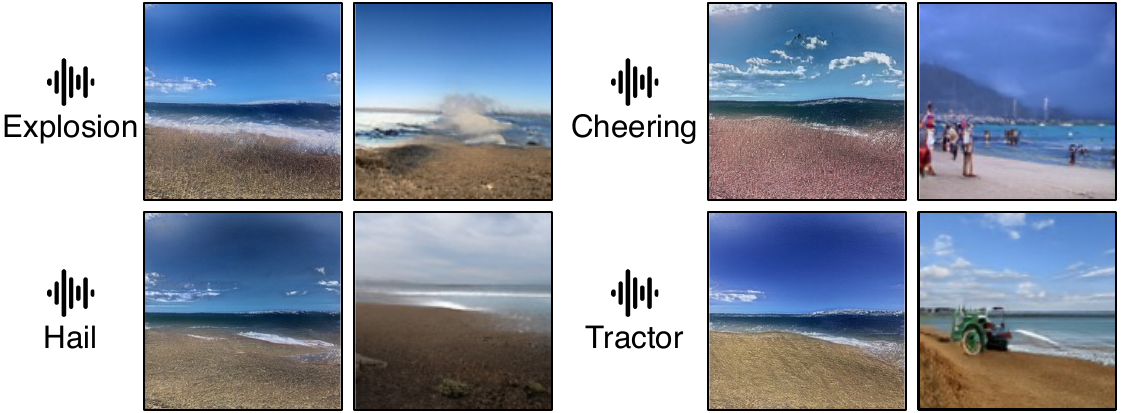}
    \blank{0.56cm} \citep{soundguide} \blank{0.90cm} Ours \blank{2.00cm}\citep{soundguide} \blank{0.90cm} Ours\blank{-0.5cm}
    
    \caption{\textbf{Qualitative comparison of our method and Lee~\emph{et al}\onedot}~\citep{soundguide}. Lee~\emph{et al}\onedot fail to insert an object on the given image with input sound. Our method, by contrast, successfully inserts objects that sound in the scene by generating a new image. Note that both works target different tasks.}
    
    \label{fig:stylize}
\end{figure}

\paragraph{Image and audio conditioned image generation}
Given an image and a semantic condition from audio, our model can generate target images in a compositional manner. This requires understanding the semantic coupling between the given image content and the audio condition. This task is similar to the recent trend of compositional image retrieval. However, here, we aim to demonstrate the compositional image generation ability of our model using audio conditions. We use simple multimodal embedding space arithmetic for image and audio-conditioned image generation.
Given an image and audio, we extract visual features $\mathbf{z^V}$ and audio features $\mathbf{z^A}$. 
We then interpolate these features in the latent space to obtain a new feature: $\mathbf{z^{new}} = \lambda\mathbf{z^V} + (1-\lambda)\mathbf{z^A}$, where $\lambda$ varies across examples. 
This new feature is fed into the image generator to synthesize an image.
As shown in~Fig.~\ref{fig:imgaud}, this simple approach effectively incorporates the sound context into the given scene, such as adding parade-looking people with marching sounds, stylizing the given building image with hail or skiing sound, and adding various vehicles on the road with respective sounds.

We further use this approach to compare our method with specifically designed the sound-guided image manipulation approach~\citep{soundguide} in~Fig.~\ref{fig:stylize}. Although this task is not explicitly targeted by our model, it emerges as a natural outcome of our design. While the results from Lee et al.~\citep{soundguide} maintain the overall content of the original image, they fail to insert objects that correspond to the sound. In contrast, our method successfully generates an image (nearly similar to the given one) by conditioning on both modalities, for instance, it adds an explosion and a tractor to the scene or makes the ocean view appear cloudy in response to hail sounds.

\begin{figure}[tp]
    \centering
    \includegraphics[width=\linewidth]{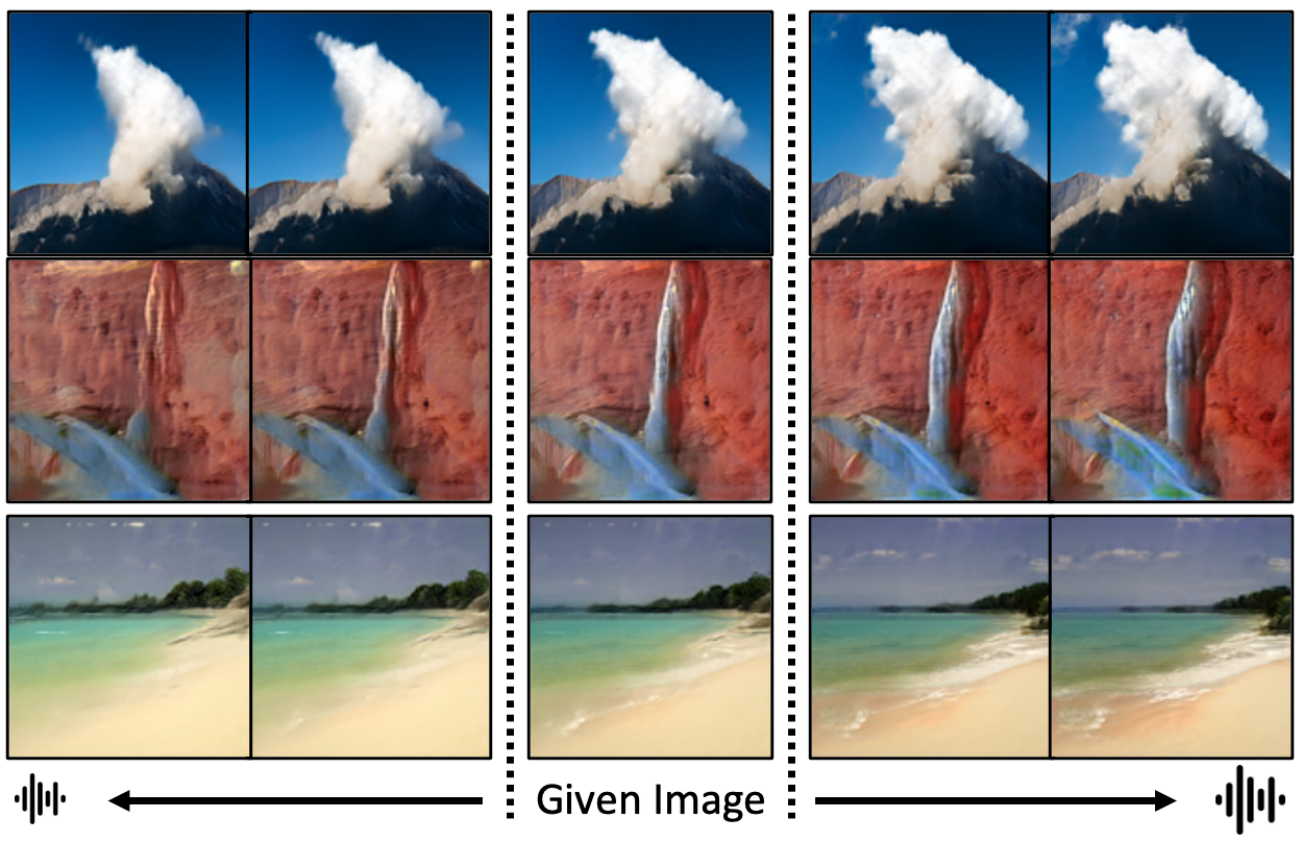}

    \caption{\textbf{Image editing by volume changes in \emph{latent space}.} We extract an image feature and noise vector by GAN inversion, and two audio features with different volumes. Then, we move the image feature in the direction of the audio feature differences.}
    \label{fig:edit}
    
\end{figure}

\paragraph{Image editing with the modifications of paired sound}
Here, we explore sound-guided image editing from a different perspective by manipulating inputs in the latent space.
Utilizing GAN inversion techniques~\citep{gan_inv1,gan_inv2}, we extract a visual feature $\mathbf{z^V_{inv}}$ and a corresponding noise vector $\mathbf{z^N_{inv}}$ for the reference image. 
We also change the volume of the associated audio to extract two distinct audio features in the embedding space, $\mathbf{z^A_1}$ and $\mathbf{z^A_2}$. 
We then adjust the visual feature by moving it in the direction of the difference between these two audio features, resulting in a new feature: $\mathbf{z^{new}} = \mathbf{z^V_{inv}} + \lambda(\mathbf{z^A_1} - \mathbf{z^A_2})$, thereby guiding the visual manipulation with audio cues. 
This new feature is fed into the image generator $G(\mathbf{z^N_{inv}}, \mathbf{z^{new}})$, enabling us to edit the original image based on corresponding sounds. As shown in Fig.~\ref{fig:edit}, simply by adjusting the volume and navigating through the latent space, we can modify visual elements such as the size of an explosion, the flow of a waterfall, or the intensity of ocean waves.

\section{Generalization on design choices}\label{generalizatinon}

As we discuss in~Sec.~\ref{sec:gap}, cross-modal transferability, which aligns audio and visual features, is the key idea behind the effectiveness of Sound2Vision. To further reiterate this and to demonstrate that our method is not dependent on any specific design choice of the model, we present several generalization results with different setups of Sound2Vision. We conduct experiments mainly from two different generalization perspectives: the architectural choice of the model and the type of audio-visual dataset used for training the model.

\begin{figure*}[tp]
    \centering
    \includegraphics[width=1\linewidth]{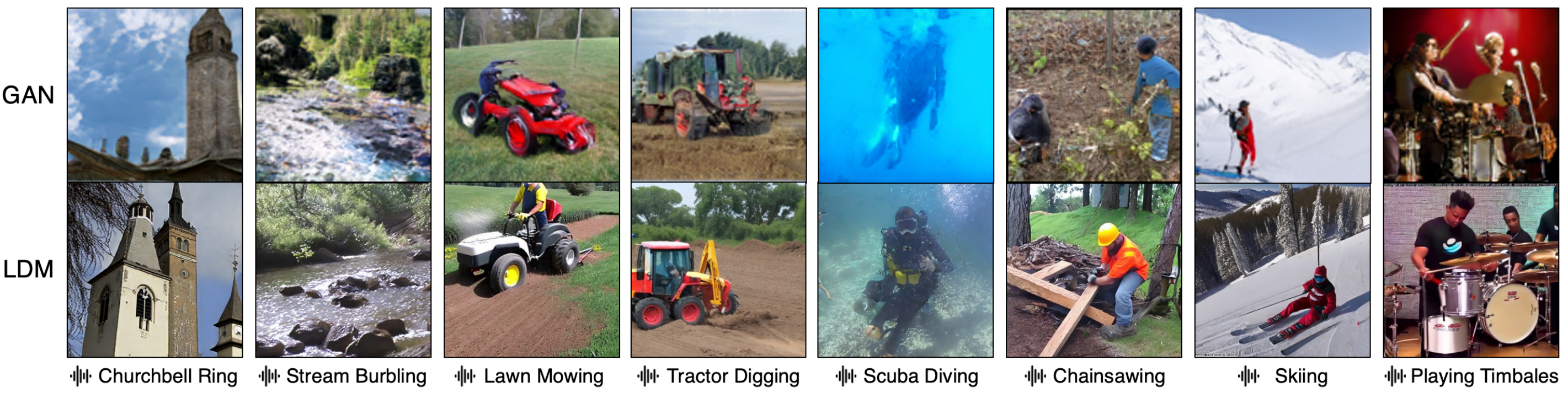}
    \caption{{\bf Qualitative results by feeding single waveform from VGGSound test set.} The first row shows results generated by using the GAN-based image generator, while the second row presents results from LDM used as an image generator.}
    \label{fig:qual_comp}
    \vspace{-4mm}
\end{figure*}

\subsection{Different architectural choice}\label{sec:dif}
Different from our original architectural choices outlined in Sec.~\ref{sec:architecure}, we switch the image generator from GAN~\citep{biggan} to the Latent Diffusion Model (LDM)~\citep{stable} and replace the image encoder from ResNet-50~\citep{he2016deep} 
to the Vision Transformer (ViT)~\citep{vit}. 
We identically follow our proposed training method as described in Sec.~\ref{ssec:training} using the VGGSound and VEGAS dataset as detailed in Sec.~\ref{sec:setup}. 
By adopting these more recent and strong components, we observe two key improvements: (1) the model produces higher quality and more realistic images, and (2) it demonstrates an ability to generate images from a broader range of sound categories than before, thereby highlighting the enhanced generative capabilities of the image generator.

\paragraph{Preliminary of Latent Diffusion Model (LDM)}
The family of diffusion models~\citep{ho2020denoising,dhariwal2021diffusion,nichol2021improved} 
aims to learn the underlying probabilistic model of the image data distribution $p(\mathbf{x})$. 
Building on these successes, LDM~\citep{stable} learns such a distribution within the latent space of the variational autoencoder, $p(\mathbf{z})$.
This involves learning the reverse Markov process over a sequence of $T$ timesteps in the latent space.
For each timestep $t=0,\dotsc,T$, the denoising function $\epsilon_{\theta}: R^d \rightarrow R^d$, where $d$ is the dimension of the latent, is trained to predict a denoised version of the perturbed $\mathbf{z_t}$ at timestep $t$, as $\epsilon_{\theta}(\mathbf{z_t},t)$. The corresponding objective can be simplify written as follows:
\begin{equation}
L_{LDM}=\mathbb{E}_{\mathbf{z_t}, t, \epsilon\in \mathcal{N}(0,I)}[\norm{\epsilon-\epsilon_\theta(\mathbf{z_t},t)}_2^2].
\end{equation}

The variant of LDM that accepts conditional inputs formulates the conditional distribution as $p(\mathbf{z}|\mathbf{c})$, where $\mathbf{c}$ serves as the conditioning input for generation.
This model integrates a cross-attention mechanism, allowing it to be conditioned with different modalities.
This is done by modifying the objective function to integrate this conditioning as:
\begin{equation}
L_{LDM}=\mathbb{E}_{\mathbf{z_t}, t, \epsilon\in \mathcal{N}(0,I)}[\norm{\epsilon-\epsilon_\theta(\mathbf{z_t},t, \mathbf{c})}_2^2]. 
\end{equation}
In our case, the LDM is initially trained to use the visual features extracted by the Vision Transformer (ViT)~\citep{vit} as the conditioning vector $\mathbf{c}$. 
Subsequently, we train the audio encoder to learn to align with these visual features, which are then used as the conditioning for sound-to-image generation.

\begin{table}[tp]
\footnotesize
\centering
\caption{\textbf{Quantitative comparison.} We compare the results of our models, using a GAN-based image generator and an LDM, and another comparison model (\cite{audiotoken}).}
\vspace{-2mm}
    \resizebox{1\linewidth}{!}{
    \begin{tabular}{ll @{\quad}cccc@{\quad}cc}
    \toprule
    &\multirow{2}{*}{\begin{tabular}[c]{@{}c@{}}Image\\ Generator\end{tabular}}&\multicolumn{4}{c}{VGGSound (50 classes)}&\multicolumn{2}{c}{VEGAS}\\
    \cmidrule(r{4mm}){3-6} \cmidrule{7-8}
     && R@1 & R@5 & FID ($\downarrow$) & IS ($\uparrow$) & R@1& R@5 \\
    \cmidrule{1-8}
    (A)&GAN & 40.71  & 77.36 & 17.97 & 19.46 & 57.44 & 84.08 \\
    (B)&LDM & \textbf{50.37}  & \textbf{81.61} & 17.23 & \textbf{19.55} & \textbf{64.40} & \textbf{85.40} \\
    (C)&AudioToken & 43.11  & 78.24 & \textbf{17.12} &16.92 & - & - \\
    \bottomrule
    \end{tabular}
    }
\label{tab:quan_gan_ldm}
\end{table}

\paragraph{Improved image quality}
Figure~\ref{fig:qual_comp} shows the qualitative comparison between images generated by GAN image generator (denoted as GAN) and those produced using the Latent Diffusion Model image generator (denoted as LDM). 
With the same input audio from various categories, the results from LDM clearly demonstrate improved image quality with more fine-grained details. This enhancement is further supported by the quantitative evaluations summarized in Table~\ref{tab:quan_gan_ldm}. Using the LDM as an image generator shows enhanced performance across all datasets and evaluation metrics. Furthermore, compared to AudioToken~\citep{audiotoken}, which are also based on LDM and trained with VGGSound, our model shows favorable results in overall metrics.

\begin{figure}[tp]
    \centering
    \includegraphics[width=\linewidth]{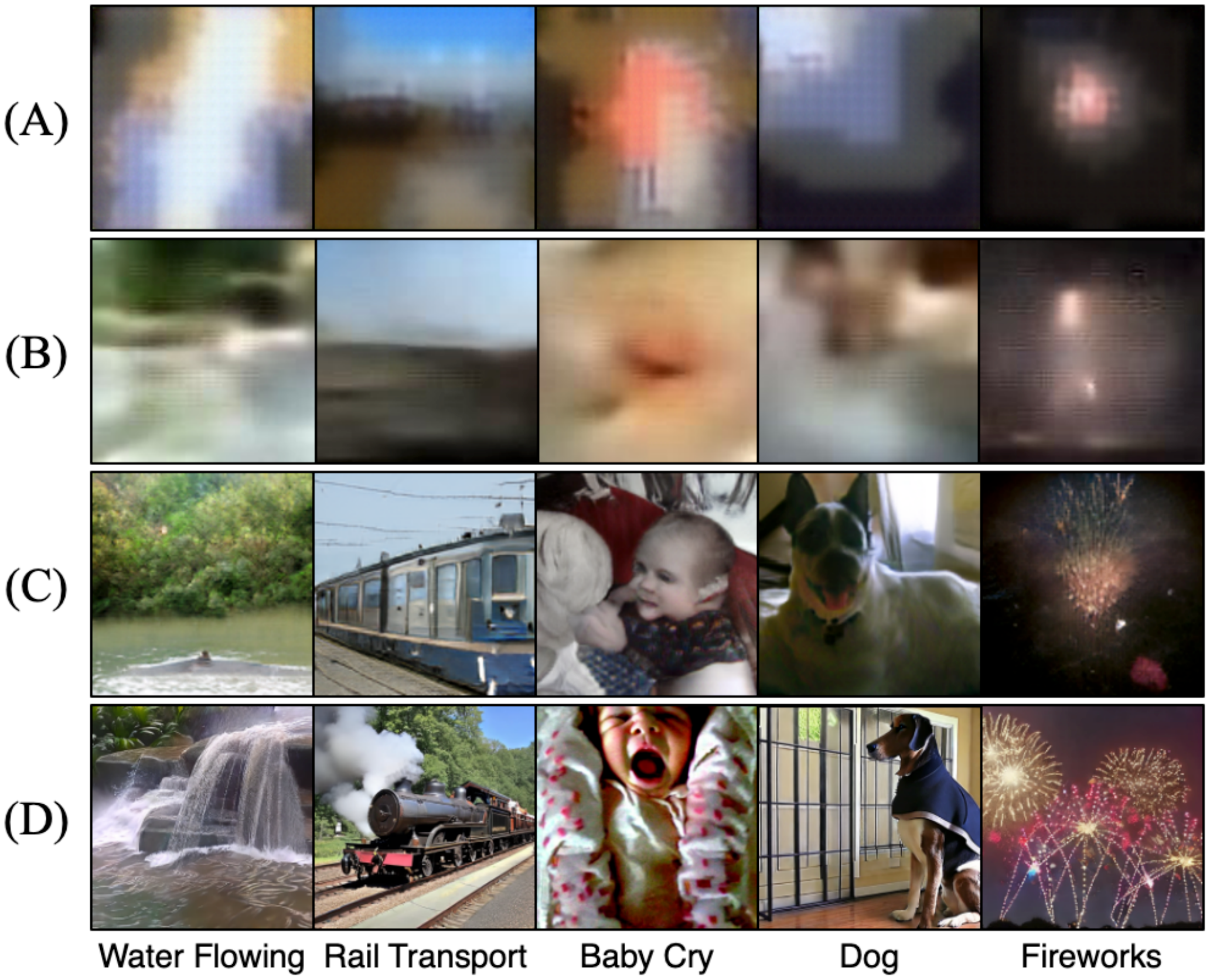}\\
    \vspace{2.5mm}
    \resizebox{0.85  \linewidth}{!}{
    \begin{tabular}{llccc}
    \toprule
    &\multirow{2}{*}{Method}&\multicolumn{3}{c}{VEGAS (5 classes)}\\
    \cmidrule{3-5}
    & & R@1 & FID ($\downarrow$) & IS ($\uparrow$)\\
    \cmidrule{1-5}
    (A)& Pedersoli~\emph{et al}\onedot~\citep{pedersoli2022estimating} & 23.10 & 118.68 & 1.19\\
    (B)& S2I~\citep{s2i} & 39.19 & 114.84 & 1.45\\
    (C)& Ours (GAN) & 77.58 & 34.68 & 4.01\\
    (D)& Ours (LDM) & \textbf{79.31} & \textbf{32.52} & \textbf{4.04}\\
    \bottomrule
    \end{tabular}
    }
        \caption{\textbf{Comparison to the existing approaches}. Our methods outperform the others both qualitatively and quantitatively in the VEGAS dataset.}
    
    \label{tab:quan_vegas}
\end{figure}

We also compare both of our models with other prior arts, S2I~\citep{s2i}\footnote{\url{https://github.com/leofanzeres/s2i}} and Pedersoli~\emph{et al}\onedot~\citep{pedersoli2022estimating}\footnote{\url{https://github.com/ubc-vision/audio_manifold}}. 
Note that Pedersoli~\emph{et al}\onedot, although not targeted for sound-to-image generation task, utilizes a VQVAE-based model~\citep{vqvae} for generating sound-to-depth or segmentation images. 
Despite our models' ability to handle a broader range of in-the-wild audio, we adopt the same training setup as S2I by training our models and Pedersoli~\emph{et al}\onedot with five categories from the VEGAS dataset for a fair comparison. 
As shown in Fig.~\ref{tab:quan_vegas}, both of our models surpass all other methods. Additionally, our proposed models generate visually plausible images, whereas previous methods often fail to produce recognizable results. 
These outcomes 
highlights
the effectiveness of our training approach, where learning visually enriched audio features combined with a robust image generator leads to superior performance.

\begin{figure}[tp]
    \centering
    \includegraphics[width=\linewidth]{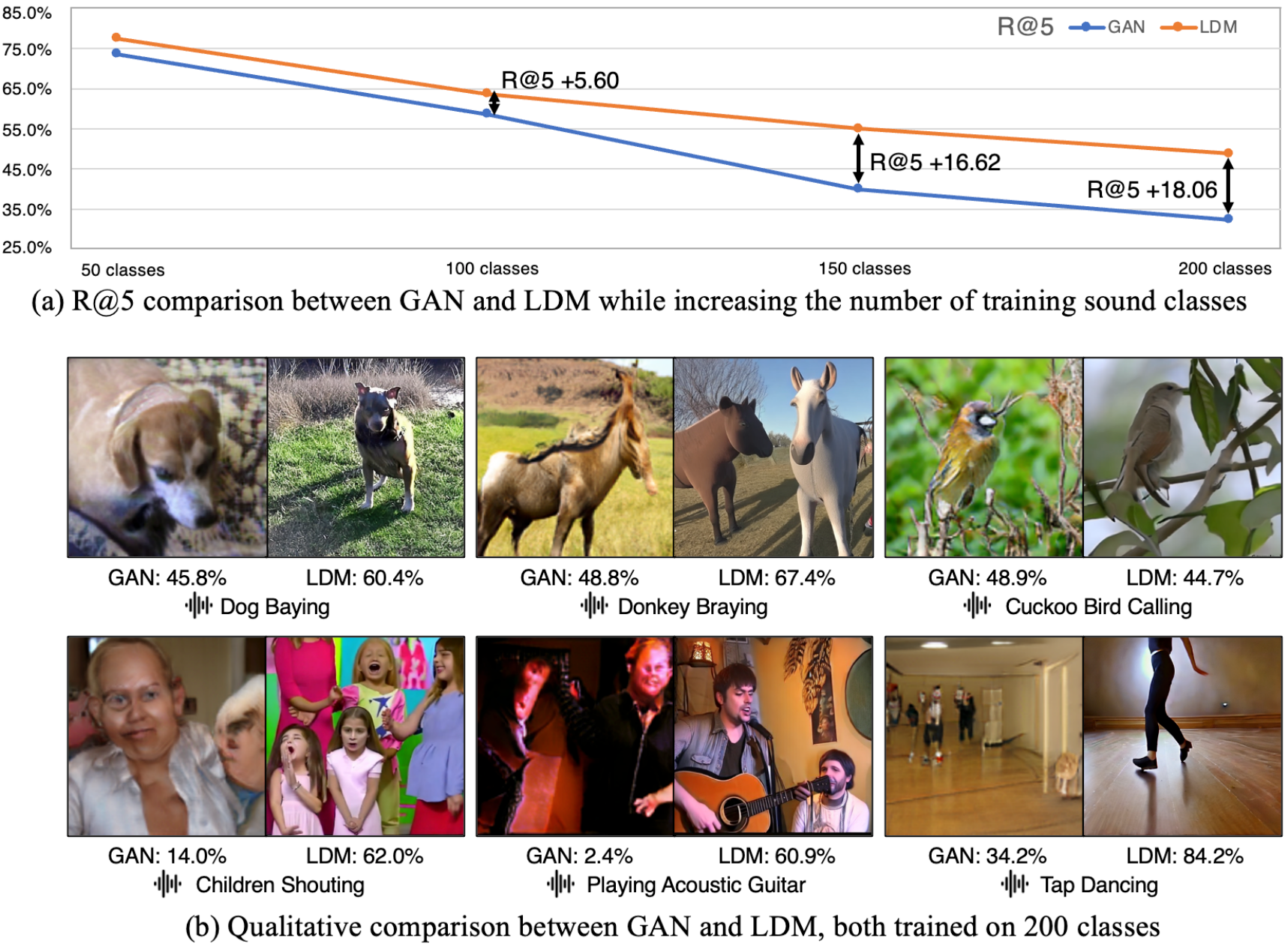}
    \caption{\textbf{Comparisons by increasing the number of training sound classes.} 
    (a) shows the CLIP retrieval score (R@5) for GAN and LDM. LDM prevents drastic degradation compared to GAN when increasing the number of training sound classes. (b) shows qualitative results for GAN and LDM, both trained with 200 sound classes.
    In animal-related classes, both GAN and LDM show favorable results. 
    In human-related classes, GAN produces unrecognizable results with poor R@5, while LDM consistently generates plausible images.}
    \label{fig:category_clip}
    
\end{figure}

\begin{figure*}[tp]
    \centering
    \includegraphics[width=1\linewidth]{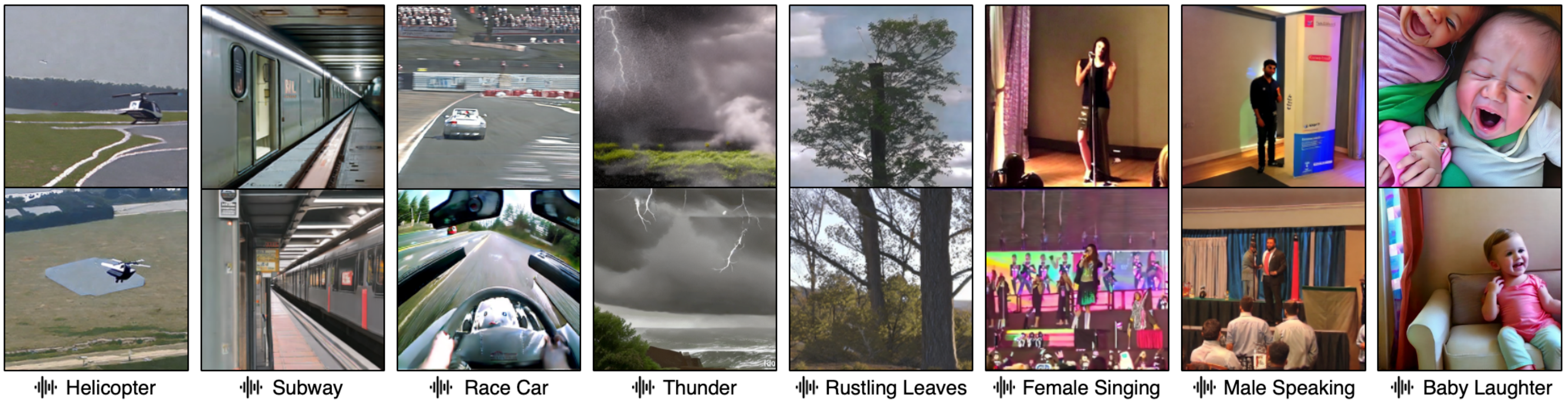}
    \caption{{\bf Qualitative results using LDM on new sound categories.} LDM can generate realistic images from diverse transportation, environmental, and human-related sounds.}
    \label{fig:qual}
    \vspace{-2mm}
\end{figure*}

\begin{figure}[tp]
    \centering
    \includegraphics[width=\linewidth]{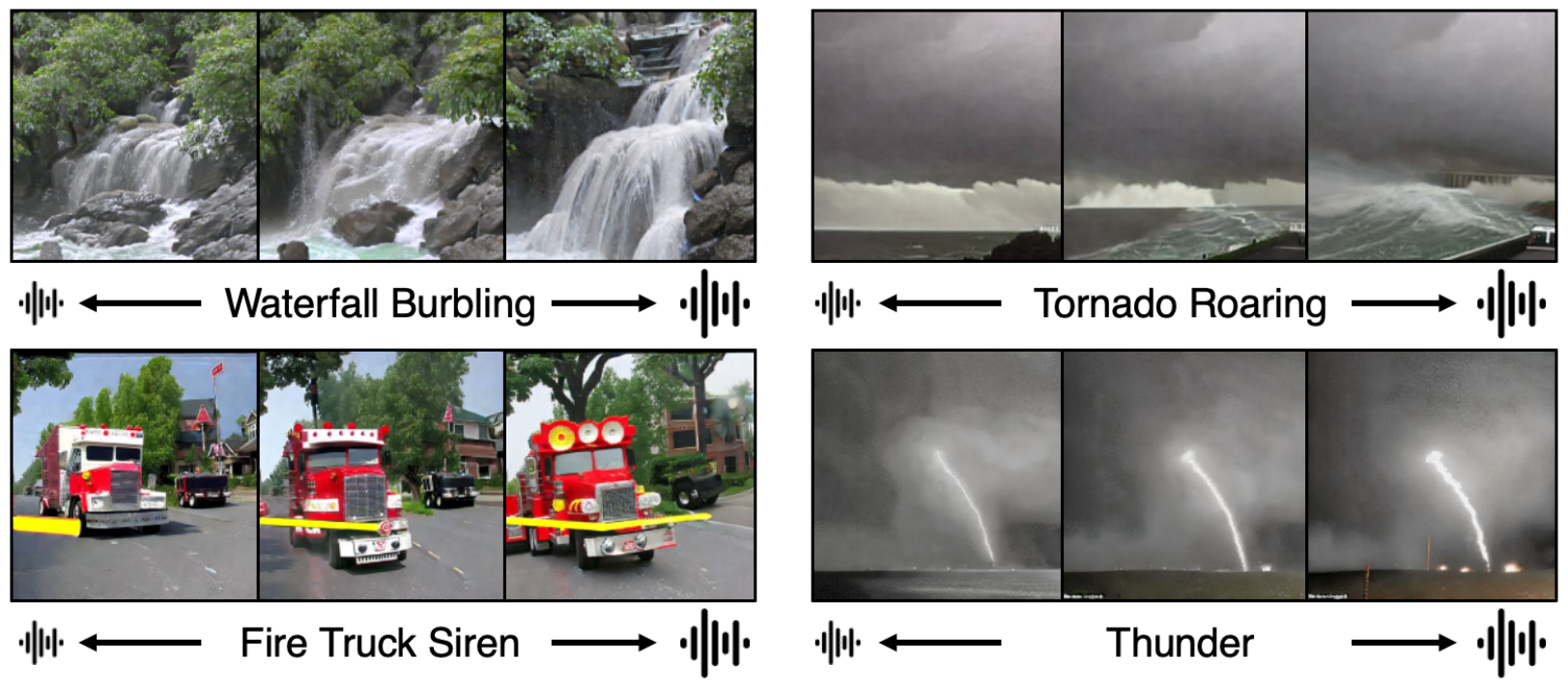}
    \caption{\textbf{Generated images with LDM by changing the volumes of the input audio.} The objects of the sound source become larger or more dynamic as the volume of the same audio increases.}
    \label{fig:volume_LDM}
    \vspace{-2mm}
\end{figure}

\paragraph{Handling more categories}
The model with the LDM image generator can generate images from a broader range of sound categories that are unattainable with the GAN-based model. 
As described in \citep{sung2023sound}, while VGGSound contains 310 categories, 50 categories among these are selected for training the model. 
The choice to use a subset of categories is because of the limitations of the GAN-based model's generative power, which struggles to produce plausible images across all categories.
Training on all categories results in degraded image quality and 
lower 
CLIP retrieval performance.
However, with the LDM model, we demonstrate that the model can handle more diverse sound categories while maintaining the image quality and reasonable range of CLIP retrieval scores.

Figure~\ref{fig:category_clip} (a) shows the CLIP retrieval R@5 results for both the GAN and LDM models, demonstrating how performance varies as the number of training sound classes increases from 50 to 200.
As shown, GAN shows a drastic performance drop in R@5, whereas the degradation in LDM follows a more gentle slope.
Interestingly, the performance of training 100 classes in GAN is similar to that of training 150-200 classes in LDM.
One of the reasons for this degradation in GAN is due to its weaker generative power, especially in human-related classes, a limitation also discussed in \citep{sung2023sound}. 
As the number of sound classes increases, the likelihood of including audio-visual pairs related to human actions also increases, resulting in poorer CLIP R@5 on GAN.

In Fig.~\ref{fig:category_clip} (b), we also provide a qualitative comparison between GAN and LDM along with CLIP R@5 results for each category, where both models are trained with 200 classes. 
In classes related to animals, both GAN and LDM produce plausible images, resulting in similar R@5 scores. 
For instance, the ``Dog Baying'' and ``Donkey Braying'' classes show approximately a 16\% degradation in CLIP R@5 from LDM to GAN, while the ``Cuckoo Bird Calling'' class shows comparable results.
However, in human action-related classes, GAN generates unrecognizable images while LDM consistently produces high-quality images with identifiable human actions.
Moreover, LDM outperforms on CLIP R@5, showing approximately a 50\% drop from LDM to GAN.

Figure~\ref{fig:qual} presents additional results across new categories that the GAN-based model could not handle. LDM can generate images from various transportations and environmental sounds, as well as human-related sounds, accurately reflecting the gender or age range.

\begin{figure}[tp]
    \centering
    \small
    \includegraphics[width=\linewidth]{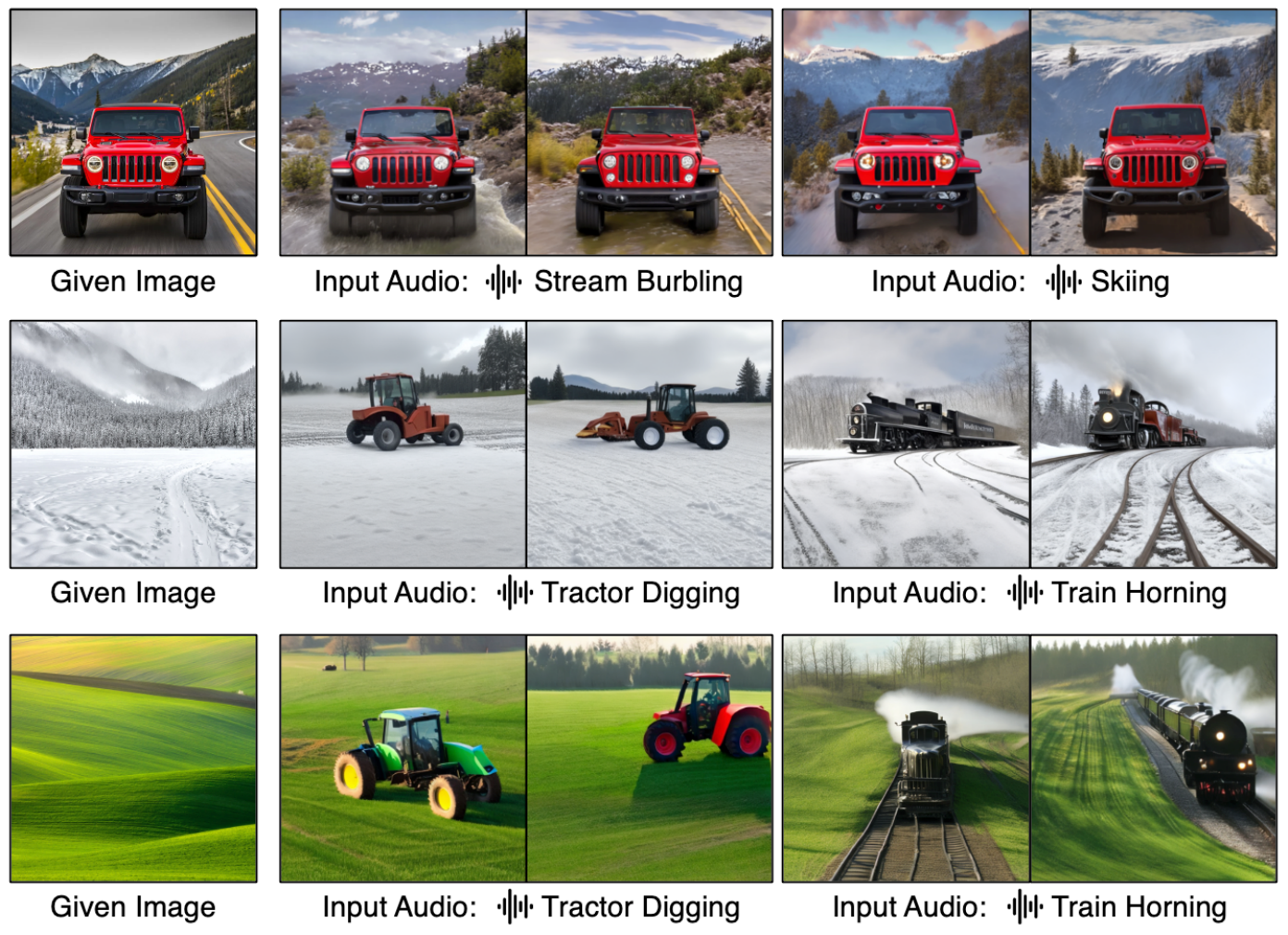}\\
    \blank{-1.4cm}(a) Inputs \blank{2.1cm} (b) Generated images
    \caption{\textbf{Generated images conditioned on image and audio with LDM.} We interpolate between a given visual feature and an audio feature in the \emph{latent space}. This interpolated feature is then fed to the LDM image generator to synthesize a novel image.}
    \label{fig:imgaud_LDM}
    \vspace{-2mm}
\end{figure}

\begin{figure*}[tp]
    \centering
    \includegraphics[width=1\linewidth]{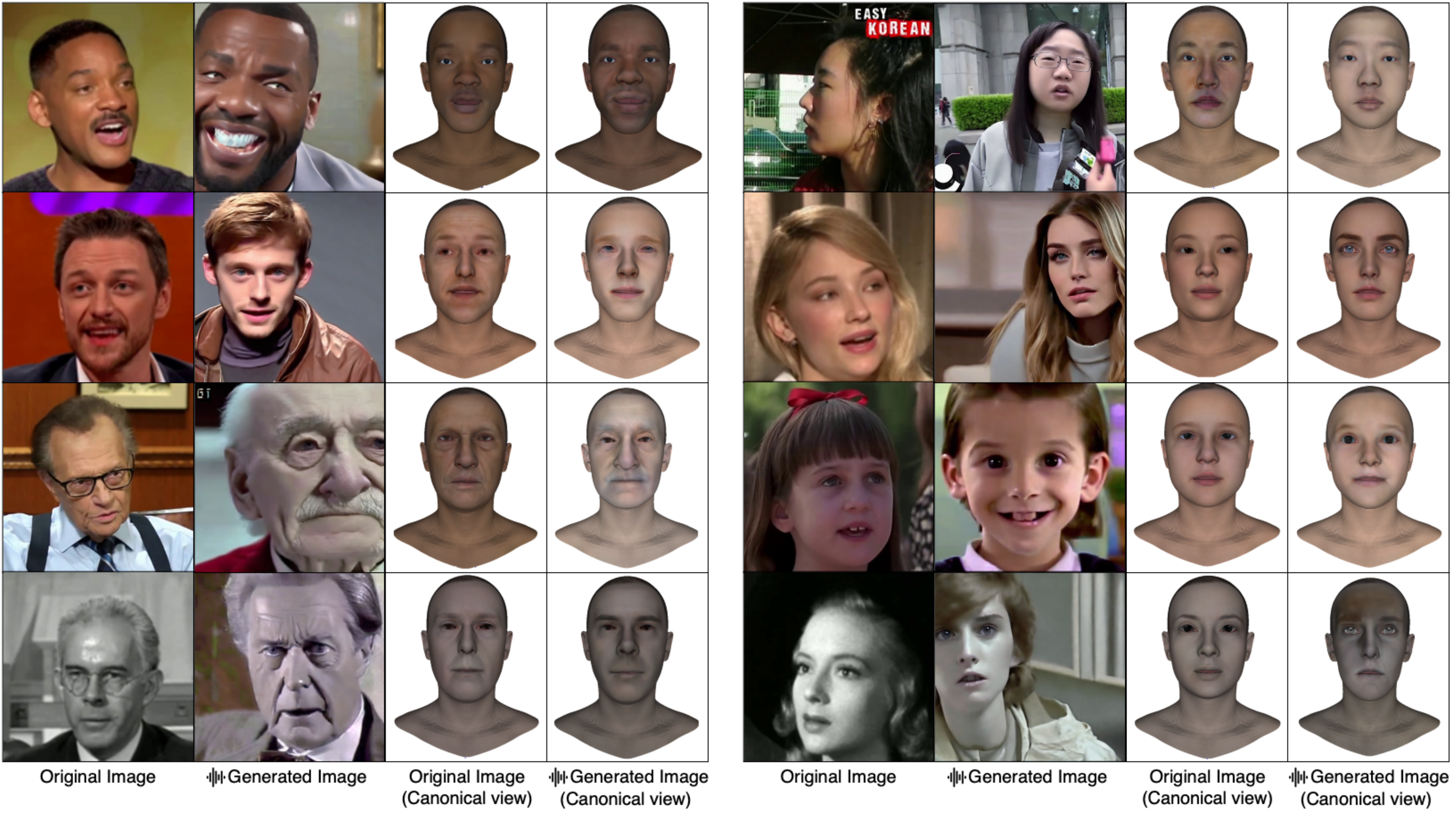}
    \caption{{\bf Qualitative results by feeding human speech from the test split of CelebV-HQ.} The left two columns display the original image from the video and the generated face images, respectively, while the third and fourth columns show the canonical 3D face mesh with textures of both the original and generated images. The model effectively learns to extract diverse information from the input speech, such as gender, ethnicity, and age range, and uses this to generate human faces.}
    \label{fig:face}
\end{figure*}

\paragraph{Controllability}
Similar to Sec.~\ref{sec:control}, the LDM version of our model also supports the capability of generating controllable outputs by manipulating inputs in both waveform and latent spaces to some extent. 
As demonstrated in Fig.~\ref{fig:volume_LDM}, the model can respond to variations in the volume of the same audio, \emph{i.e.}, manipulation in the waveform space.
For instance, increasing the volume results in a more forceful waterfall and a larger depiction of thunder.
Figure~\ref{fig:imgaud_LDM} illustrates the results of image and audio conditioned image generation, which involves manipulation in the latent space. 
We specifically refer to \citep{liu2022compositional} and employ compositional generation techniques to incorporate multiple concepts from both image and audio. 
As shown, the model successfully manipulates the given car image with several environmental sounds, such as a burbling stream and skiing sound.
In other examples, the model successfully inserts sound sources into the given images. One clear observation, as also presented above, is that the quality of the generated images is higher than that of the GAN version (see Fig.~\ref{fig:imgaud}). Additionally, it maintains the integrity of the given image better than the GAN version while manipulating it with the sound. However, this is primarily due to the superior image generation ability of the LDM. As all the examples and results in this section show, regardless of the image generator used, the alignment of the cross-modal signal is the key factor in achieving these capabilities, and our learning objective facilitates this.

\subsection{Different dataset type}\label{sec:dataset}
While the audio-visual pairs constructed from the VGGSound~\citep{vggsound} and VEGAS~\citep{vegas} dataset mostly contain environmental and in-the-wild events, \emph{i.e.}, generic sounds, we demonstrate that our training method can be effectively applied to different types of audio-visual datasets, including speech and faces. 
Although there is no direct one-to-one mapping from speech to face, Speech2Face~\citep{speech2face} has shown that models can learn to extract facial information from speech and generate faces that resemble the original speaker.
Specifically, we utilize the CelebV-HQ~\citep{zhu2022celebv} dataset, which includes 35,666 video segments featuring a diverse range of talking faces from various identities, ages, genders, and appearances. 
We arbitrarily select 20,000 videos from which we extract speech and image frames to construct our training dataset, while the remaining videos are used for testing the model. Using this dataset, the training scheme identically follows Sec.~\ref{sec:dif}. 
The audio encoder is trained to align with the visual features extracted by a Vision Transformer (ViT), which is compatible with the LDM. 
After training, the speech inputs are processed by the audio encoder and then fed into the LDM to generate facial images from the speech.

\begin{figure*}[tp]
    \centering
    \includegraphics[width=\linewidth]{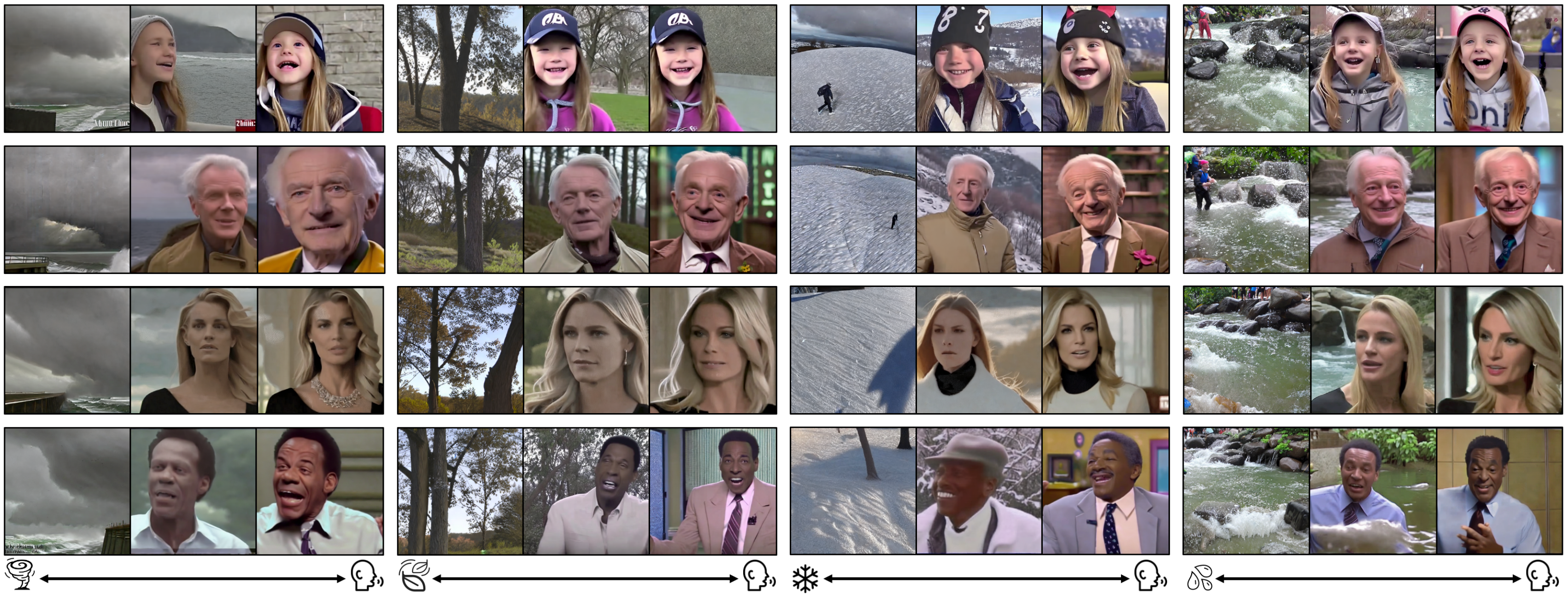}
    \caption{\textbf{Generated images conditioned on both speech and environmental sound.} 
    We interpolate between two audio features in the latent space — one derived from speech and the other from environmental sound. This interpolated feature is then input into the LDM to integrate both concepts into a novel image.
    \humanEmoji, \tornadoEmoji, \treeEmoji, \snowEmoji, and \waterEmoji\,\,denote speech, tornado roaring, leaves rustling, footsteps on snow, and splashing water sound, respectively.}
    \label{fig:speech_ambient}
    \vspace{-2mm}
\end{figure*}

\paragraph{Results on CelebV-HQ test samples}
Figure~\ref{fig:face} shows faces generated from diverse types of input speech.
Interestingly, without using explicit complex modeling to learn the relationship between speech and facial attributes, the model effectively distills rich visual information into audio features. It successfully extracts facial attributes from the input speech and accurately reflects them in the generated images, including ethnicity (first row), gender (second row), and age range (third row).
Moreover, the model captures subtle cues from the speech, such as a vintage audio effect from old black-and-white films, and incorporates this cue by generating faces with grayish colors, mimicking the style of old black-and-white movies, as shown in the last row.

Since the original and generated faces have diverse head poses and are captured from different camera viewpoints, we facilitate comparison by reconstructing these into canonical 3D meshes, displayed in the third and fourth columns.
Specifically, we feed both the original and generated images into MICA~\citep{zielonka2022towards} to reconstruct 3D faces and use FFHQ-UV~\citep{bai2023ffhq} to apply textures to the reconstructed faces. This visualization method allows us to clearly observe that the model can effectively generate faces from speech, resembling the appearances similar to the original faces.

\paragraph{Combining speech and environment sound}
Having trained one audio encoder on environmental in-the-wild sounds as described in Sec.~\ref{sec:dif} and another on a speech dataset as in Sec.~\ref{sec:dataset}, we are able to generate images that reflect both environmental and speech sounds in a single image. Given an audio feature from environmental sounds, $\mathbf{z^A_E}$, and another from speech, $\mathbf{z^A_S}$, we can interpolate between these features to define a composite audio feature, $\mathbf{z^{new}} = \lambda\mathbf{z^A_E} + (1-\lambda)\mathbf{z^A_S}$, where $\lambda$ varies across examples. 
This new audio feature is then fed into the LDM to generate images that incorporate both types of content. 
Figure~\ref{fig:speech_ambient} shows the images that integrate both signals into one image. 
The leftmost image is generated solely from environmental sound, the rightmost image from speech, and the center image reflects both interpolated signals. 
Interestingly, this simple interpolation in the latent space enables the placement of humans in diverse environments, such as snowy or cloudy scenes, while preserving their identity.

\section{Conclusion}\label{conclusion}
In this paper, we propose Sound2Vision, a model designed to generate images relevant to given audio inputs. 
This task presents inherent challenges due to the significant information gap between audio and visual signals as audio lacks visual information, and audio-visual pairs do not always correspond directly. 
Previous approaches have been constrained by the limited number of sound categories they can generate and the low quality of the images produced.
Our method addresses these challenges by enriching audio features with visual knowledge and selecting well-correlated audio-visual pairs for training, thereby successfully producing richly detailed images with diverse characteristics.
Moreover, our model offers controllability over inputs, allowing for more creative outcomes compared to previous methods.
Additionally, we analyze the geometric properties of the multimodal embeddings, demonstrating how our learning approach effectively aligns audio-visual signals for cross-modal generation, which is the key concept behind our work. 
This analysis shows that our method is agnostic to specific design choices, highlighting its generalizability through successful integration with various model architecture choices and types of audio-visual data.

\noindent\textbf{Data availability statements.}
All data supporting the findings of this study are available online.
The VGGSound dataset can be downloaded from \url{https://www.robots.ox.ac.uk/~vgg/data/vggsound/}. 
The VEGAS dataset can be downloaded from \url{https://github.com/postech-ami/Sound2Scene}.
The CelebV-HQ dataset can be downloaded from \url{https://github.com/CelebV-HQ/CelebV-HQ}.

{\small
\bibliographystyle{spbasic}
\bibliography{refs}
}

\end{document}